\documentclass{article}
\pdfoutput=1  

\usepackage{cite} 
\usepackage{graphicx}
\usepackage{amssymb}

\newcommand{\eq}[1]{Eq.~\ref{eq.#1}} 
\newcommand{\eqbare}[1]{\ref{eq.#1}} 

\newcommand{\fig}[1]{Fig.~\ref{fig.#1}}

\newcommand{\tbl}[1]{Table~\ref{table.#1}}
\newcommand{\tbls}[1]{Tables~\ref{table.#1}} 
\newcommand{\tblbare}[1]{\ref{table.#1}}

\newcommand{\sect}[1]{Section~\ref{sect.#1}}

\newcommand{\sectlabel}[1]{\label{sect.#1}}
\newcommand{\eqlabel}[1]{\label{eq.#1}}
\newcommand{\figlabel}[1]{\label{fig.#1}}
\newcommand{\tbllabel}[1]{\label{table.#1}}

\newcommand{\figwidth}{4in}
\newcommand{\figwidthS}{2.5in} 

\newcommand{\invivo}{\textit{in vivo}}

\newcommand{\scenario}[1]{{\sc #1}}

\newcommand{\BoltzmannConstant}{\ensuremath{k_B}}

\newcommand{\density}{\ensuremath{\rho}}
\newcommand{\viscosity}{\ensuremath{\eta}}
\newcommand{\viscosityKinematic}{\ensuremath{\nu}}  
\newcommand{\Tbody}{\ensuremath{T_{\mbox{\scriptsize body}}}}
\newcommand{\alphaRMS}{\ensuremath{\alpha_{\mbox{\scriptsize rms}}}} 

\newcommand{\Pdrag}{\ensuremath{P_{\mbox{\scriptsize drag}}}}
\newcommand{\Ppropel}{\ensuremath{P_{\mbox{\scriptsize propel}}}}
\newcommand{\Pinternal}{\ensuremath{P_{\mbox{\scriptsize internal}}}}
\newcommand{\Pfriction}{\ensuremath{P_{\mbox{\scriptsize friction}}}}
\newcommand{\kfriction}{\ensuremath{k_{\mbox{\scriptsize friction}}}}

\newcommand{\Spropulsion}{\ensuremath{S_{\mbox{\scriptsize p}}}}
\newcommand{\Vnonpropulsion}{\ensuremath{V_{\mbox{\scriptsize np}}}}
\newcommand{\Snonpropulsion}{\ensuremath{S_{\mbox{\scriptsize np}}}}
\newcommand{\VnonpropulsionMin}{\ensuremath{\Vnonpropulsion^{\mbox{\scriptsize min}}}}
\newcommand{\SnonpropulsionMin}{\ensuremath{\Snonpropulsion^{\mbox{\scriptsize min}}}}

\newcommand{\aSphere}{\ensuremath{a_{\mbox{\scriptsize sphere}}}}

\newcommand{\Dmotile}{\ensuremath{D_{\mbox{\scriptsize m}}}}

\newcommand{\meter}{\mbox{m}}
\newcommand{\centimeter}{\mbox{cm}}
\newcommand{\millimeter}{\mbox{mm}}
\newcommand{\micron}{\mbox{$\mu$m}}
\newcommand{\nanometer}{\mbox{nm}}
\newcommand{\hour}{\mbox{hr}}
\newcommand{\second}{\mbox{s}}
\newcommand{\millisecond}{\mbox{ms}}

\newcommand{\radian}{\mbox{rad}}

\newcommand{\MHz}{\mbox{MHz}}
\newcommand{\kHz}{\mbox{kHz}}
\newcommand{\kg}{\mbox{kg}}

\newcommand{\picowatt}{\mbox{pW}}

\newcommand{\Kelvin}{\mbox{K}}
\newcommand{\Pascal}{\mbox{Pa}}
\newcommand{\gigapascal}{\mbox{GPa}}
\newcommand{\piconewton}{\mbox{pN}}
\newcommand{\volt}{\mbox{V}}
\newcommand{\ms}{\meter^2/\second}

\title{Using Surface-Motions for Locomotion of Microscopic Robots in Viscous Fluids}
\author{Tad Hogg\\Institute for Molecular Manufacturing\\Palo Alto, CA}

\begin{document}
\maketitle

\begin{abstract}

Microscopic robots could perform tasks with high spatial precision, such as acting in biological tissues on the scale of individual cells, provided they can reach precise locations.
This paper evaluates the feasibility of \invivo\ locomotion for micron-size robots. 
Two appealing methods rely only on surface motions: steady tangential motion and small amplitude oscillations. These methods contrast with common microorganism propulsion based on flagella or cilia, which are more likely to damage nearby cells if used by robots made of stiff materials. 
The power potentially available to robots in tissue supports speeds ranging from one to hundreds of microns per second, over the range of viscosities found in biological tissue. 
We discuss design trade-offs among propulsion method, speed, power, shear forces and robot shape, and relate those choices to robot task requirements. This study shows that realizing such locomotion requires substantial improvements in fabrication capabilities and material properties over current technology.

\end{abstract}

\section{Introduction}

Robots with sizes comparable to bacteria could be useful for many biological research and medical applications~\cite{freitas98,freitas06,martel07,hill08}.
Such robots could act at locations specified to precision comparable to the size of individual cells, i.e., several microns. These locations may only be recognizable once the robot is within several microns of the site, e.g., particular receptors on cell surfaces. For such precise positioning, the robots will need to find and move to their target locations themselves. Thus autonomous locomotion is a key capability for microscopic robots.

Locomotion of microscopic robots faces two major challenges. 
The first is identifying methods appropriate for the robots' physical environment. At these sizes, viscous forces and Brownian motion are significant~\cite{dusenbery09}. Thus the physics of microfluidics~\cite{fung97,happel83,kim05,squires05,white05} require different locomotion methods than larger robots~\cite{purcell77}.

The second major challenge is fabricating the propulsion components and assembling them into complete robots. 
Some fabrication techniques have been demonstrated for small robots in fluids. 
One example is propulsion by magnetic fields~\cite{dreyfus05,abbott09},
which can move microrobots containing
ferromagnetic particles through blood
vessels~\cite{ishiyama02,martel07a,olamaei10}. 
Other demonstrated micromachines use flagellar motors~\cite{behkam07,martel08,zhang09} and cilia~\cite{zhou08}.
However, locomotion of these machines is externally controlled, e.g., via magnetic fields or changing chemicals in the machine's environment, and thus do not provide autonomous locomotion.

Microorganisms use various locomotion mechanisms. A common method is moving extended structures, such as flagella and cilia~\cite{brennen77,lauga09,guasto12,jahn72,purcell77}.
These examples and current micromachines suggest flagella or cilia would be useful for robots.
However,  such appendages present significant fabrication and operational challenges for \invivo\  operation near cells and other robots.

The fabrication challenges for appendages include creating the active internal structures of flagella and cilia and assembling many cilia on the robot surface. 
A significant operational challenge is the potential for cutting nearby cells and tangling with nearby robots. Moreover, appendages expose large surface area to the environment, which could lead to fouling or immune reaction.
Reliably avoiding these events significantly increases the complexity of the robot controller. If a robot needs to shut down (e.g., due to component failure), appendages could become tangled by subsequent passive motion in fluids or movement of nearby cells.

To avoid these challenges, this paper focuses on the feasibility of propulsion via motion of the robot's surface without extended appendages.
While not as commonly studied as propulsion by flagella or cilia, some microorganisms move without using appendages~\cite{ehlers96,ehlers11a,leshansky07,keeley04}.
Two such methods are examined in this paper: steady surface motion that is flush with the robot surface and small-amplitude surface oscillations. 

This paper is a theoretical study of design trade-offs for robots that cannot yet be manufactured. It extends previous models of microorganism locomotion by examining non-locomotion design constraints and potential implementations of these models for robots. These implementations allow estimating power dissipation inside the robots, which is usually neglected in studies of microscopic locomotion.

The next section describes locomotion requirements and prototypical scenarios for microscopic robots in biological fluids. Using these scenarios as examples, the following sections evaluate the two propulsion methods. 
\sect{safety} examines the safety of these methods.
The following two sections discuss the effects of robot shape and Brownian motion on locomotion.
\sect{design choice} combines these discussions to describe design trade-offs among the propulsion methods. The final section summarizes the results and suggests directions for further study.

\section{Locomotion Performance and Scenarios}

This section discusses performance goals for locomotion, relevant tissue properties, and example scenarios.

\subsection{Performance Measures and Constraints}
\sectlabel{performance}

Locomotion involves several performance measures~\cite{brennen77}, and trade-offs among them. 
These include speed, maneuverability, and propulsive force. 
The maximum force depends on the strength of robot actuators and will generally be larger than the force required to move at the nominal locomotion speed.

Self-propelled robots will not necessarily need to move over long distances. Injection or passive flow could get the robots within a few hundred microns of most cells in the body. Or larger devices could carry the robots to the vicinity of their operating locations, which is particularly relevant for tasks involving machines operating cooperatively at various size scales in the same region (e.g., individual cells and the tissues they form)~\cite{hogg05}.
In such cases, autonomous locomotion need only enable robots to move about $100\,\micron$ in a few minutes, which is a speed of about $1\,\micron/\second$. 
Similar speeds arise in moving through protective mucus layers to underlying tissue before the mucus layer is shed, carrying away any embedded robots. Specifically, mucus layers range in thickness from tens to hundreds of microns and turn over in tens of minutes to hours~\cite{lai09}.

Robots must move around obstacles. 
For example, a cell is about $10\,\micron$ in size, so following a cell surface requires orientation changes of about $90^\circ$ after $10\,\micron$ of movement. This takes about 10 seconds when moving at $1\,\micron/\second$, i.e., an angular rotation rate of about $10^\circ/\second$.

These modest speed requirements leave ample design scope for other criteria, especially conservative designs emphasizing reliability and safety over maximizing performance.
Higher speeds extend the range of applications. For instance, blood flows in small vessels at speeds up to about $1\,\millimeter/\second$~\cite{freitas99}.
Robots able to move faster than the flow could travel upstream, e.g., to track chemical gradients extending downstream from a source on the vessel wall~\cite{hogg06a}.
Tasks needing higher speed even if possibly producing minor damage to tissue, e.g., responding to acute injury~\cite{freitas00}, are beyond the scope of this paper.

The limited resources available to robots constrain their locomotion.
One constraint is power, which robots may obtain in various ways~\cite{freitas99,soong00,wang07}. 
For instance, isolated micron-size robots may obtain tens to hundreds of picowatts from oxygen and glucose in tissue~\cite{hogg10,freitas99}, though such power generators have not yet been experimentally realized.
If the available power is insufficient for steady motion, the robot could move in bursts, accumulating fuel between these bursts or multiplexing movement with other tasks.

Another constraint is the robot's surface area.
Locomotion requires exerting forces against the fluid. In the absence of extended appendages, these forces must arise from motion of the robot surface. Propulsion mechanisms must share the surface with other components, such as sensors and pumps.  

Safe locomotion requires that the robots not significantly disturb nearby cells. 
Disturbance can be direct, e.g., due to collisions.
The low speeds considered in this paper make collision damage to cells unlikely~\cite{freitas99}. 
Another direct effect is from stiff flagella cutting cells, which the methods discussed in this paper avoid by not using extended appendages. 
Disturbance also arises indirectly from shear forces propagated through the fluid.
A measure of this disturbance is the magnitude of fluid speed vs.~distance from a moving robot. Smaller values mean less shear force on nearby cells, less drag from nearby walls and less hydrodynamic interaction among nearby robots, which could simplify controls.

\subsection{Thrust Force and Power Dissipation}
\sectlabel{power}

Locomotion has two contributions to power dissipation.
First is the locomotion power, $\Ppropel$, dissipated through viscosity in the fluid. Second is the dissipation, $\Pinternal$, within the robot to actuate the propulsion mechanism. 
The sum $\Ppropel+\Pinternal$ gives the power directly required for locomotion. Additional losses within the robot, such as to control the motion or to generate and distribute power~\cite{freitas99} are not considered here.

External power use arises from the fluid flow around the moving object.
A simple case is an externally forced spherical robot. 
Specifically, Stoke's law gives the force required to drag a sphere of radius $a$ at speed $U$ through an unbounded fluid of viscosity $\viscosity$
\begin{equation}\eqlabel{Stokes law}
F = 6\pi \viscosity a U
\end{equation}
when the Reynolds number, $a U \density/\viscosity$, is small, where \density\ is the density of the fluid. The power dissipated in the fluid is $\Pdrag=F U$.
Conversely, for an arbitrary propulsion mechanism that moves the sphere at speed $U$, applying an external force $F$ to the robot against the direction of the motion will stop the robot. Thus \eq{Stokes law} estimates the thrust force of a propulsion mechanism capable of moving at speed $U$.

A common measure of locomotion effectiveness is its \emph{hydrodynamic efficiency}~\cite{lighthill52,lauga09}: the ratio $\Pdrag/\Ppropel$ of the power required for an external force to drag the object through the fluid to the power required for the object to propel itself at the same speed. 
Hydrodynamic efficiency is typically a few percent~\cite{purcell77}.

Studies of microorganism locomotion typically focus entirely on hydrodynamic efficiency. A full evaluation also accounts for power dissipated inside the robot. This is particularly important when hydrodynamic efficiency is large~\cite{leshansky07,spagnolie10} so most of the power use arises from internal dissipation.

Internal dissipation depends on the details of the actuator mechanisms.
Various effects contribute to friction in micromachines~\cite{krim02,vanossi13}, depending on the microstructure of the surfaces, lubrication, applied forces and operating speeds.
Robots made of precisely structured, stiff materials could, in theory, be much more efficient~\cite{drexler92} than current micromachines, in part due to avoiding viscous losses inside the robot by excluding fluid from the interior.
In this case, internal dissipation is primarily due to friction between moving actuators and their housings. 
Smooth nanoscale surfaces separated by atomic-scale distances (i.e., a few tenths of a nanometer) moving past each other at speed $v$ well below that of sound in the material have friction dominated by phonon scattering~\cite{drexler92}. This leads to power loss
\begin{equation}\eqlabel{friction}
\Pfriction = \kfriction S v^2
\end{equation}
where $S$ is the area of the moving surfaces and $\kfriction$ is a constant depending on the materials and their spacing. For stiff materials, $\kfriction$ is generally less than $1000\,\kg/(\meter^2 \second)$~\cite{drexler92,freitas99}. We use this value for an upper bound estimate of internal power loss, i.e., taking $\Pinternal \approx \Pfriction$.
Thus, unlike the complexity of friction in micromachines with poorly defined geometry at nanometer scales, friction dissipation in smooth nanoscale surfaces has a simple dependence on area and speed. This allows estimating internal power dissipation for the implementations considered in this paper more readily than is possible for current micromachines.

For comparison, the viscous friction for flat surfaces moving past each other at speed $v$ while separated by a layer of fluid with viscosity $\viscosity$ and thickness $d$ is $\viscosity S v/d$. This dissipates power $(\viscosity/d) S v^2$, so $\kfriction=\viscosity/d$. Extrapolating to distances of a few tenths of a nanometer gives $\kfriction \approx 10^7\,\kg/(\meter^2 \second)$ for fluids with viscosity similar to water.
Similarly, measurements of drag for micromachine rotors~\cite{chan11} gives $\kfriction$ of about this size.
More precisely defined micron-scale surfaces can have lower friction~\cite{yang13}, but still significantly larger than the value of $\kfriction$ given above.
Thus the internal dissipation we consider, while achievable in principle, is several orders of magnitude smaller that of current micromachines. This lower friction is important for the feasibility of the propulsion mechanisms considered here, which might otherwise require more power than available to robots.

\subsection{Scenarios}

Fluid density and viscosity determine the nature of locomotion. In biological tissues, viscosity varies by orders of magnitude while density is roughly the same. Thus as example scenarios we consider a wide range of viscosity but constant density.
For the  short distances we consider ($\sim100\micron$, about 5 to 10 cell diameters), we assume a homogeneous fluid.

Specifically, these scenarios are moving in 1) a low viscosity fluid (comparable to water) at about $100\,\micron/\second$, and 2) a high viscosity fluid $10^4$ times more viscous than water, which is a typical value for mucus or cell cytoplasm~\cite{freitas99}.
Since $\Ppropel$ is proportional to the product of viscosity and the square of locomotion speed, we consider a speed of $1\,\micron/\second$ in the high viscosity scenario to give the same value of $\Ppropel$ for both scenarios.

Medical nanorobots could vary in size, from about one to tens of microns. Mobile robots able to access individual cells will generally be small, e.g., to allow passing through the smallest blood vessels and between or into cells. We focus on such sizes, i.e., robots that are a few microns in size.
Robot shape also influences locomotion behavior. We mainly focus on spherical robots for simplicity.

\begin{table}
\centering
\begin{tabular}{lclcc}
scenario				&	& both &\scenario{low}		& \scenario{high} \\
\hline 
speed of sound			& $c $		&$1500 \,\meter/\second$ \\
density				& $\density$	&$1000 \,\kg/\meter^3$ \\
ambient temperature    	&   $\Tbody$	&$310\,\Kelvin$   \\
viscosity	& $\viscosity$		& & $10^{-3}\,\Pascal \cdot \second$	&  $10\,\Pascal \cdot \second$\\
kinematic viscosity		& $\viscosityKinematic =\viscosity/\density$ & & $10^{-6}\,\ms$ & $10^{-2}\,\ms$\\
\hline 
radius of spherical robot	& $a$			& $1\,\micron$ \\
locomotion speed	& $U$	& & $100\,\micron/\second$	& $1\,\micron/\second$ \\
\hline 
Reynolds number	&$a U/\viscosityKinematic$	& & $10^{-4}$	& $10^{-10}$ \\
force    			& $F = 6\pi \viscosity a U$ & & $2\,\piconewton$		& $200\,\piconewton$ \\
power to drag  		& $\Pdrag = F U$ & & $2\times 10^{-4}\,\picowatt$	& $2\times 10^{-4}\,\picowatt$ \\
\end{tabular}
\caption{Scenarios corresponding to motion in low and high viscosity biological fluids. The \scenario{low} scenario corresponds to water and the \scenario{high} scenario to fluids $10^4$ times more viscous than water. Values in the third column apply to both scenarios.}\tbllabel{scenarios}
\end{table}

\tbl{scenarios} gives the fluid and robot parameters for these scenarios. In both cases, the Reynolds number is much less than one, indicating viscous forces dominate the fluid behavior~\cite{purcell77}.

\section{Propulsion by Steady Tangential Surface Motion}
\sectlabel{treadmill}

Steady tangential movement of the robot surface involves motion whose speed is constant in time but may vary at different locations on the surface. Sustaining this motion requires transporting surface material inside the robot to return the material to its original location on the surface.
For instance, \fig{constant motion}a illustrates axially symmetric motion on a sphere where material moves from north to south pole, expanding and contracting as it moves. Material accumulating at the south pole moves inside the robot to reappear at the north pole.

\begin{figure}[th]
\centering
\begin{tabular}{cc}
\includegraphics[width=\figwidthS]{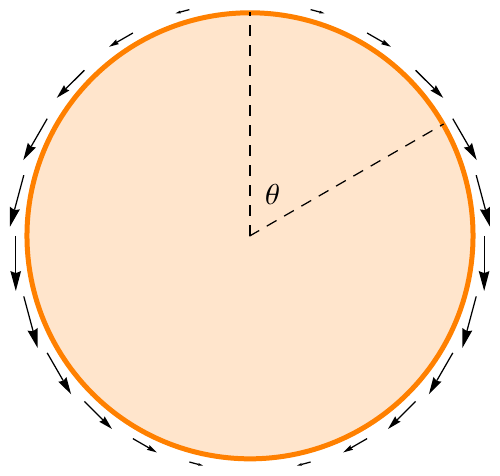} &
 \includegraphics[width=\figwidthS]{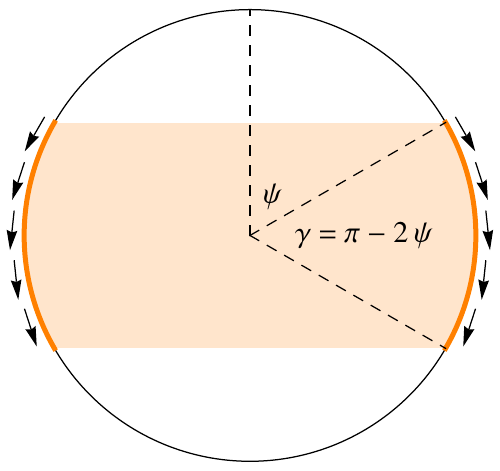} \\
(a) & (b) \\
\end{tabular}
\caption{Axially-symmetric tangential motion, indicated by arrows, on a cross section of a sphere. The shaded areas indicate the extent of the motion. Points travel inside the robot to return to their original locations (not shown).
(a) Speed at polar angle $\theta$ proportional to $\sin \theta$. 
(b) Constant speed in a band, of angular size $\gamma$, around the equator, consisting of $\theta$ in the range $\psi \le \theta \le \pi-\psi$ where $\psi=(\pi-\gamma)/2$, and no propulsion on the rest of the surface. 
The vertical axis is the direction of locomotion.}\figlabel{constant motion}
\end{figure}

A robot needs some surface area for other components, such as sensors. 
Embedding them in a moving surface significantly increases the difficulty of fabricating the components, including their connection to other parts of the robot. Thus practical propulsion with this surface motion uses only part of the surface, in contrast with studies allowing motion on the full surface~\cite{leshansky07,michelin10,osterman11}.
\fig{constant motion}b shows such an approach: the surface only moves within a band of angle $\gamma$ centered on the equator, and, for simplicity, the motion is with constant velocity.

\subsection{Performance}

For simplicity in quantifying the performance of steady surface motion, we focus on spherical robots. This analysis applies to motion at low Reynolds number and generalizes to spheroids~\cite{leshansky07}.
For a sphere with radius $a$, tangential surface motion $\mathbf{u}(\theta,\phi)$ at the point specified by spherical coordinates $\theta,\phi$ gives locomotion velocity $\mathbf{U}$ and angular velocity $\mathbf{\Omega}$~\cite{stone96b} 
\begin{eqnarray}
\mathbf{U}		&=& -\frac{1}{4\pi a^2} \int_S \mathbf{u} \;dS \eqlabel{tangential motion} \\
\mathbf{\Omega}	&=&	 -\frac{3}{8\pi a^3} \int_S \mathbf{n\times u} \;dS \eqlabel{tangential rotation}
\end{eqnarray}
where $\mathbf{n}$ is the unit outward normal vector to the sphere surface $S$. For axisymmetric surface motion the sphere does not rotate, i.e., $\mathbf{\Omega}=0$, and the power required to produce this locomotion is~\cite{stone96b} 
\begin{equation}\eqlabel{tangential power}
\Ppropel =  \frac{2 \viscosity}{a} \int_S \mathbf{|u|}^2 \;dS
\end{equation}

The surface speed $u(\theta)=v \sin(\theta)$ directed longitudinally  gives the maximum hydrodynamic efficiency, of $50\%$, for a sphere~\cite{leshansky07,osterman11}. In this expression, $v$ is the maximum surface speed and $U=\frac{2}{3} v$ from \eq{tangential motion}.
Tangential motion using only part of the surface, as shown in \fig{constant motion}b, is only somewhat less efficient, as shown in \tbl{tangential performance}.

For a robust design, the surface speed $v$ will be well within the strength limits of the robot's actuators. Thus the maximum surface speed the actuators can produce, and hence thrust available to the robot, as determined by the strength of the surface material and motors internal to the robot, could be considerably larger than the values indicated in the table.

\begin{table}
\centering
\begin{tabular}{llll}
scenario				&	&\scenario{low}		& \scenario{high} \\
propulsion band angle	& $\gamma$ 		& $60^\circ$	& $60^\circ$ \\
surface area fraction		& $\sin(\gamma/2)$	&  $1/2$ 	&  $1/2$\\
surface speed			& $v$			& $210\,\micron/\second$	& $2.1\,\micron/\second$\\
locomotion speed		& $U=\frac{1}{4} v (\gamma +\sin (\gamma ))$		& $100\,\micron/\second$	& $1\,\micron/\second$\\
power				& $8 \pi  a \viscosity  v^2 \sin (\gamma/2)$	& $0.00055\,\picowatt$	& $0.00055\,\picowatt$\\
hydrodynamic efficiency	& $\frac{3}{64} (\gamma +\sin (\gamma ))^2 \csc (\gamma/2)$		& $0.34$	& $0.34$\\
max. thrust			& $6\pi a \viscosity U$	& $1.9\,\piconewton$	& $190\,\piconewton$ \\
\end{tabular}
\caption{Performance of tangential motion in an equatorial band on a sphere, illustrated in   \fig{constant motion}b, computed from \eq{tangential motion} and \eqbare{tangential power} using parameters of \tbl{scenarios}.}\tbllabel{tangential performance}
\end{table}

\begin{figure}
\centering \includegraphics[width=\figwidthS]{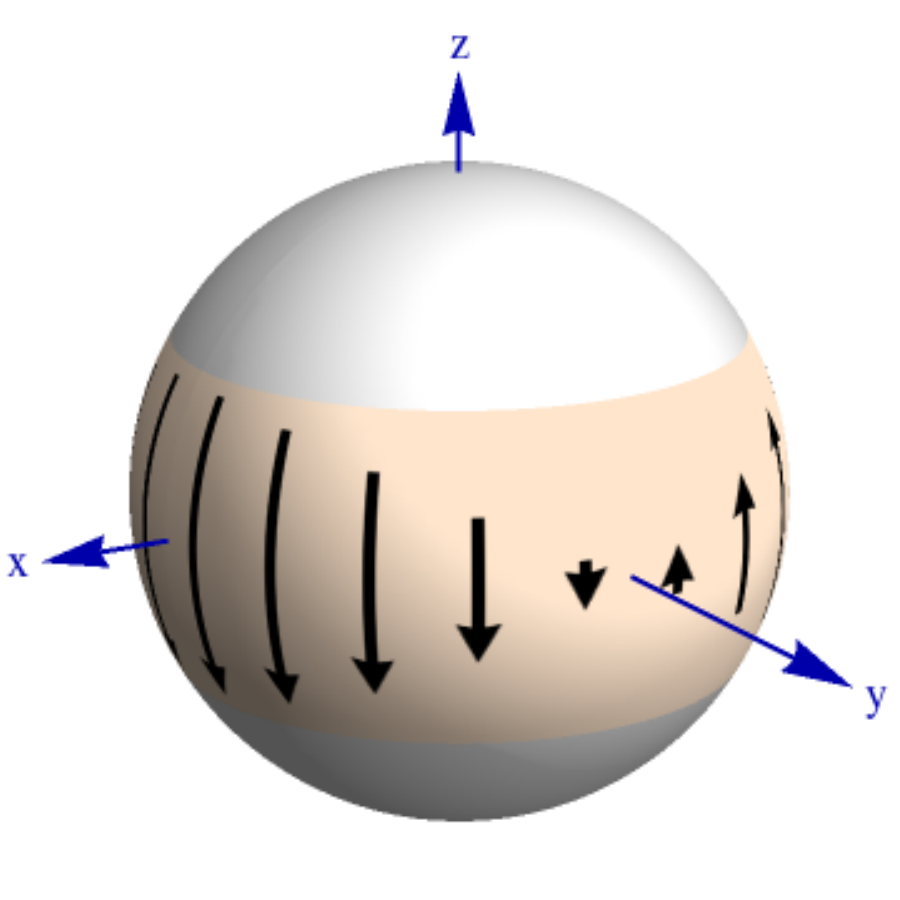}
\caption{Surface motion $u(\theta,\phi)=v \cos(\phi)$ on an equatorial band that rotates the sphere around the $y$-axis. Arrows indicate the direction of surface motion, with lengths proportional to the speed.}\figlabel{rotation}
\end{figure}

Rotating the robot requires non-axially symmetric surface motions.
An example is longitudinally-directed speed $u(\theta,\phi)=v \cos(\phi)$ on an equatorial band of angle $\gamma$, shown in \fig{rotation}. In this case, \eq{tangential rotation} gives the angular velocity shown in \tbl{tangential performance-rotation}, which is oriented along the negative y-axis.
The time required to change orientation by angle $\alpha$ is $t=\alpha/\Omega$. If the actuators can move in both directions through the equatorial band, there is no need to rotate the robot to move in the opposite direction. In that case, the largest required rotation is $90^\circ$, requiring time $t=\pi/(2\Omega)$. For the example in \tbl{tangential performance-rotation}, this rotation takes about $15\,\millisecond$. By comparison, a robot moving at $100\,\micron/\second$ travels a distance of a cell diameter in about $100\,\millisecond$, so this rotation rate readily allows adjusting direction without significantly affecting the time required for the motion around cells.

\begin{table}
\centering
\begin{tabular}{lll}
propulsion band angle	& $\gamma$ 		& $60^\circ$ \\
surface area fraction		& $\sin(\gamma/2)$	&  $1/2$ \\
max. surface speed		& $v$			& $267\,\micron/\second$\\
angular velocity		& $\Omega = \frac{3}{4} \frac{v}{a} \sin(\gamma/2)$		& $100\,\radian/\second$\\
\end{tabular}
\caption{Example performance for rotation via tangential motion on a sphere with radius $a=1\,\micron$.}\tbllabel{tangential performance-rotation}
\end{table}

\subsection{Implementation}
\sectlabel{tangential-implementation}

Two implementations that approximate steady tangential motions are treadmills and small wheels on the surface. These do not produce exactly tangential motion, e.g., where the tread penetrates the robot surface, nor precisely the expressions for the surface speed, $u(\theta,\phi)$, discussed above, e.g., due to gaps between wheels. Nevertheless, these mechanisms approximately implement the locomotion performance described above while also providing estimates of actuator capabilities and internal power dissipation as described in this section. Moreover, some microorganisms employ locomotion mechanisms similar to the treadmill behavior discussed here~\cite{keeley04,menard01}.

Theoretical studies indicate small electrostatic motors could actuate the treadmills and wheels~\cite{drexler92,freitas99}. As with the other components for these robots, fabricating such motors is a major technical challenge.

\subsubsection{Treadmills}

\fig{treadmill} shows treadmills on an equatorial band, corresponding to the parameters of \tbl{tangential performance}, i.e., propulsion using half the sphere's surface area. In this diagram, the tread runs over the outer surface, thereby producing motion of a spherical surface evaluated here\footnote{An alternate implementation has the tread running straight between the two bearings, corresponding to a robot shaped as a cylinder with spherical caps at each end.}.
The performance analysis given above corresponds to negligible gaps between treadmills, which thus cover most of the equatorial band.

\begin{figure}
\centering
\begin{tabular}{cc}
\includegraphics[width=\figwidthS]{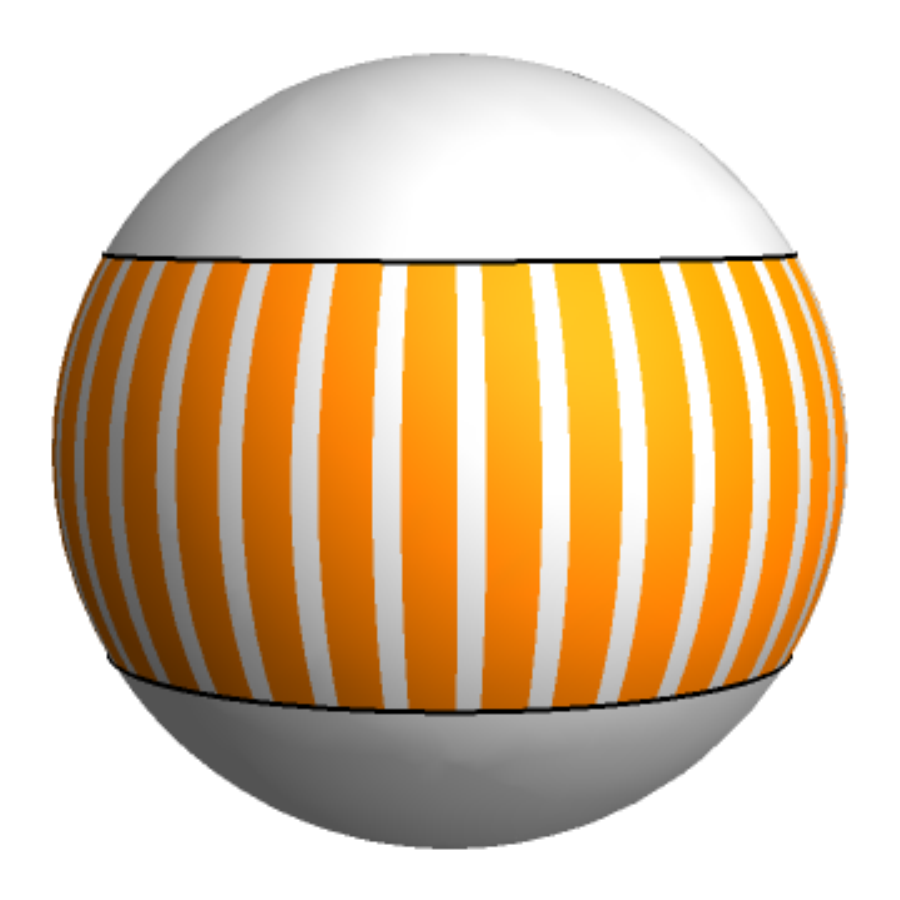} &
 \includegraphics[width=\figwidthS]{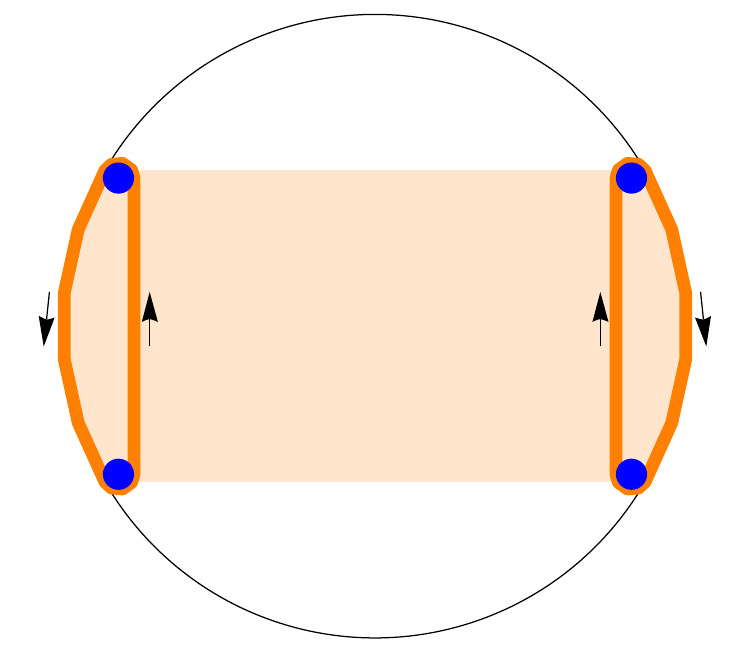} \\
(a) & (b) \\
\end{tabular}
\caption{(a) Schematic of a set of treadmills on a band around the equator of a sphere. Treads move vertically, from top to bottom of the band and are separated by narrow gaps. (b) Cross section showing internal movement of treadmills. Blue disks indicate the bearings, and arrows indicate motion of the treadmills when they all move in the same direction, giving axially symmetric motion. The vertical axis is the direction of locomotion.}\figlabel{treadmill}
\end{figure}

To evaluate structural requirements for a treadmill, consider a tread of length $L$, width $W$, area $A=L W$ and thickness $h$ pulled at speed $v$ around bearings of radius $r$. The bearings turn at angular frequency $\omega=v/r$.
To estimate of the force applied to the fluid, consider a treadmill moving a distance $d$ from a wall (where fluid speed is zero). The force on the tread area exposed to the fluid is $F=\viscosity \nabla v A$ with velocity gradient $\nabla v\approx v/d$,
This force stretches the tread, resulting in strain $(F/(W h))/E$ where $W h$ is the tread's cross section area and $E$ is Young's modulus of the material.
For bending the tread around the bearings, the distance of the tread along inner and outer edges is $\pi r$ and $\pi(r+h)$, respectively, giving strain of $h/(r+h/2)\approx h/r$ for $h\ll r$ and corresponding stress of  $E h/r$.

\begin{table}[htdp]
\begin{center}
\begin{tabular}{lll}
\multicolumn{3}{c}{tread}\\
width	& $W$	&$100\,\nanometer$\\
length exposed to fluid	& $L$	&$1\,\micron$\\
thickness	& $h$	&$1\,\nanometer$\\
Young's modulus	& $E$	&$1000\,\gigapascal$\\
tread speed	& $v$	&$210\,\micron/\second$\\
\hline \multicolumn{3}{c}{bearing}\\
radius	& $r$	&$50\,\nanometer$\\
rotation rate	& $f=v/(2\pi r)$	&$0.7\,\kHz$\\
angular velocity	& $\omega=v/r$	& $4200\,\radian/\second$\\
\hline \multicolumn{3}{c}{tread bent around bearing}\\
strain		& $h/r $	&$ 2\%$\\
stress		& $h E/r$	&$20 \,\gigapascal$\\
\hline \multicolumn{3}{c}{forces when moving $d=100\,\nanometer$ from a wall}\\
drag on tread from fluid	& $F=\viscosity (v/d) L W$	&$ 0.2\,\piconewton$\\
tension on tread cross section	& $ F/(h W)$	&$2000\,\Pascal$\\
\end{tabular}
\end{center}
\caption{Example parameters for treadmill. Tread speed corresponds to the \scenario{low} scenario of \tbl{tangential performance}.}
\tbllabel{treadmill example}
\end{table}

The tread tension due to fluid drag in \tbl{treadmill example} is far below the failure strength, typically at least $10^{10}\,\Pascal$, of strong materials. So such treadmills can readily provide forces that produce the speeds considered here, even in fluids with several orders of magnitude higher viscosity.
The tread has substantial strain as it bends around the bearing. If necessary to reduce this strain, the bearing radius could be somewhat larger, thereby requiring more of the robot's interior volume for propulsion components. Alternatively, the tread could be thinner. An extreme limit for a thin tread is a single-atom layer, e.g., graphene with elastic modulus $E=10^{12}\,\Pascal$,  and breaking strength around $10^{11}\,\Pascal$~\cite{lee08}.
In summary, \tbl{treadmill example} shows that strong materials could implement tangential motion with treadmills.

\subsubsection{Wheels}

The speed along a treadmill is constant, so treadmills extending the full width of the equatorial band (as shown in \fig{treadmill}) have speed, $u(\theta,\phi)$, that does not vary with the polar angle $\theta$. This is suitable for the constant speed discussed in \tbl{tangential performance}, but not for motions such as illustrated in \fig{constant motion}a.
Using a series of shorter treadmills could approximate such motions. An alternative is to forego the tread and instead use a set of closely-spaced wheels. 
If the wheels have small diameter compared to the size of the robot, their motion approximates the tangential surface motion discussed above\footnote{This contrasts with propulsion by a few larger thin wheels~\cite{freitas99} which extend a significant distance from the robot surface and hence could cut into nearby cells.}.
\fig{wheels} illustrates this implementation.

\begin{figure}
\centering \includegraphics[width=\figwidthS]{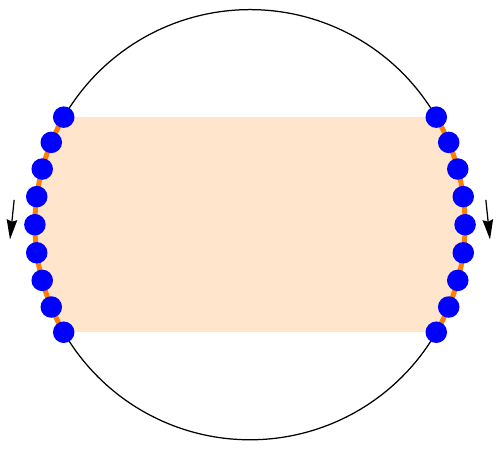}
\caption{Schematic wheel propulsion in a band around the equator, shown on a cross section of the sphere. Blue disks indicate the wheels, and arrows indicate motion when the wheels all move in the same direction, giving axially symmetric motion. In this case, wheels on the left (right) rotate counterclockwise (clockwise).  The vertical axis is the direction of locomotion.}\figlabel{wheels}
\end{figure}

The stress on the wheels due to their rotation is negligible compared to their strength. Specifically, the stress on a disk of density $\rho$ and radius $r$ rotating with angular velocity $\omega$ is $\sim \rho v^2$ where $v=r \omega$ is the velocity of the outer edge~\cite{freitas99}.  
As an example, wheels similar to the bearings of \tbl{treadmill example} have $r=50\,\nanometer$, $v\sim 200\,\micron/\second$, and density of $\sim5000\,\kg/\meter^3$. This gives stress $\rho v^2 \approx 10^{-4}\,\Pascal$, far below the material strength.

\subsubsection{Internal power dissipation}

\begin{figure}
\centering
\scalebox{0.95}{
\begin{tabular}{cc}
\includegraphics[width=\figwidth]{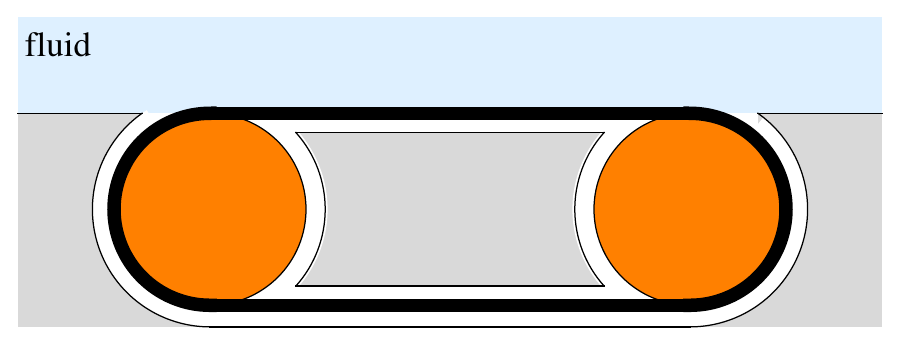} &
 \includegraphics[width=1.8in]{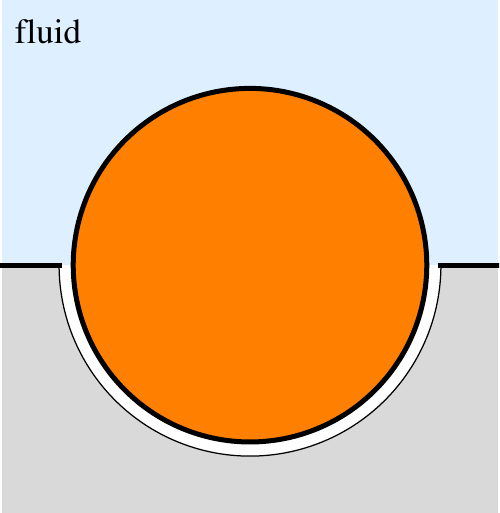} \\
(a) & (b) \\
\end{tabular}
}
\caption{(a) Schematic treadmill and bearings sliding in a housing (gray). The bearing and gap sizes are exaggerated relative to length of the tread. (b) Schematic wheel positioned halfway inside the robot (gray). The gap size is exaggerated relative to wheel. In both cases, the rotating parts also slide against robot housing on either side, into and out of the page (not shown), which separates these components from those at neighboring meridians of  longitude on the robot surface.}\figlabel{tangential-internal}
\end{figure}

For the tread parameters of \tbl{treadmill example}, about 50 treadmills cover the equatorial band of the sphere (\fig{treadmill}a) with total sliding area of about $20\,\micron^2$ for all the treads and their bearings (\fig{tangential-internal}a). With this sliding surface area and the tread speeds matching the surface speeds in \tbl{tangential performance}, \eq{friction} gives $\Pfriction$ of $10^{-3}\,\picowatt$ and $10^{-8}\,\picowatt$ for the \scenario{low} and \scenario{high} scenarios, respectively.
Comparing to \tbl{tangential performance}, these estimates suggest internal power dissipation is a bit larger than the external loss for the \scenario{low} scenario, similar to the situation for microorganisms using cilia~\cite{keller77}.
On the other hand, $\Pfriction$ is much lower than external loss for the \scenario{high} scenario, so the high viscosity of the fluid dominates the total power use.

For wheels using the same width and radius as for the treadmill bearings of \tbl{treadmill example}, about 500 wheels cover the equatorial band on the sphere, with surface area comparable to that of the treadmill internal surfaces. 
The position of the wheel center relative to the robot surface is a design choice. For example, with the center at the robot surface (\fig{tangential-internal}b), half of each wheel's circumference slides against the robot housing, giving $\Pfriction$ of $5\times 10^{-4}\,\picowatt$ and $5 \times 10^{-9}\,\picowatt$ for the \scenario{low} and \scenario{high} scenarios, respectively. If wheel centers are positioned a bit below the robot surface, more of their surface will slide against the robot housing, the smaller portion in the fluid will more closely approximate tangential motion, and there will be larger gaps between the portions of neighboring wheels that move next to fluid.

With the approximate nature of these estimates for internal power dissipation, these values indicate treadmills and wheels have similar internal dissipation.

\subsubsection{Comparing treadmills and wheels}

Both treadmill and wheel implementations approximately produce tangential surface motion.

Compared to wheels, treadmills have fewer bearings and fewer breaks in the robot surface. Each such break might leak fluid into the robot,  so minimizing the number of breaks could increase locomotion reliability. 

Treadmills approximate tangential surface motion more accurately than wheels. In particular, the narrow gaps between adjacent wheels along a meridian of longitude will have large fluid shear where the wheels' edges enter and exit the robot. Moreover, some area on the sides of the wheels extends into the fluid, producing additional drag. Thus wheels will produce more fluid dissipation than treadmills, especially at higher speeds, e.g., for the \scenario{low} scenario.
With tread bearings completely below the robot surface, as illustrated in \fig{tangential-internal}a, and fluid excluded from the robot interior, the tread motion in the fluid is nearly tangential, even near the bearings. Thus the large gradient in surface velocity near the bearings is close to that at the edges of the equatorial band in the idealized model (\fig{constant motion}b). The match to purely tangential motion can be improved with bearings further inside the robot combined with a shallow-sloped channel guiding the tread to the surface.

On the other hand, wheels can produce a wider range of surface motions since the wheels along a longitude meridian can move at different speeds. For example, varying the wheel speed as a function of latitude can improve hydrodynamic efficiency compared to constant speed vs.~latitude of treadmills. Moreover, force sensors on each wheel give more spatial resolution about the robot surface than the force on a tread, thereby enabling higher resolution feedback control.

\section{Propulsion by Surface Oscillations}
\sectlabel{oscillation}

Periodic surface oscillations can propel a robot via traveling surface waves~\cite{brennen74}. These waves can travel over the full extent of the surface, e.g., from north to south pole on a sphere, even though individual points on the surface move only a small distance. 

To specify the oscillation, we label points on the surface by their location on the undistorted sphere. For axially symmetric motion, all points with the same polar angle move the same way, so a point's polar angle on the undistorted sphere, $\vartheta$, completely determines its motion. 
\fig{oscillating surface} illustrates this notation.
A useful expression for the position of the point at time $t$ is a sum over oscillation modes:
\begin{eqnarray}\eqlabel{distorted sphere}
r &=& a \left( 1 + \epsilon \sum_{n=2}^\infty \alpha_n(t) P_n(\cos \vartheta) \right) \\
\theta &=& \vartheta + \epsilon \sum_{n=2}^\infty \beta_n(t) V_n(\cos \vartheta)
\end{eqnarray}
where amplitudes $\alpha_n$ and $\beta_n$ are periodic functions with period $T=2\pi/\omega$,  $\omega$ is the angular oscillation frequency, $P_n(x)$ is the $n^{th}$ Legendre polynomial and
\begin{equation}
V_n(x) = \frac{1}{n+1} P_n^1(x)
\end{equation}
where $P_n^1$ is the associated Legendre polynomial of order 1~\cite{oliver10}.  
The quantity $\epsilon$ provides a convenient overall scale for the oscillation while the relatives sizes of amplitudes $\alpha_n$ and $\beta_n$ specify the shape of the oscillation.
Modes with large $n$ correspond to surface oscillations with short wavelengths.

We define the amplitudes at time $t$ as
\begin{eqnarray}\eqlabel{amplitudes}
\alpha_n(t) &=& \Re(e^{-i \tau +i \gamma_n}) A_n \\
\beta_n(t)  &=& \Re(e^{-i \tau + i \eta_n}) B_n \nonumber
\end{eqnarray}
where $\Re(z)$ is the real part of the complex number $z$,  $\tau = \omega t$, $A_n$ and $B_n$ are nonnegative, and the angles $\gamma_n$ and $\eta_n$ specify the phases of the radial and tangential components of the motion, respectively.

\begin{figure}[th]
\centering \includegraphics[width=\figwidth]{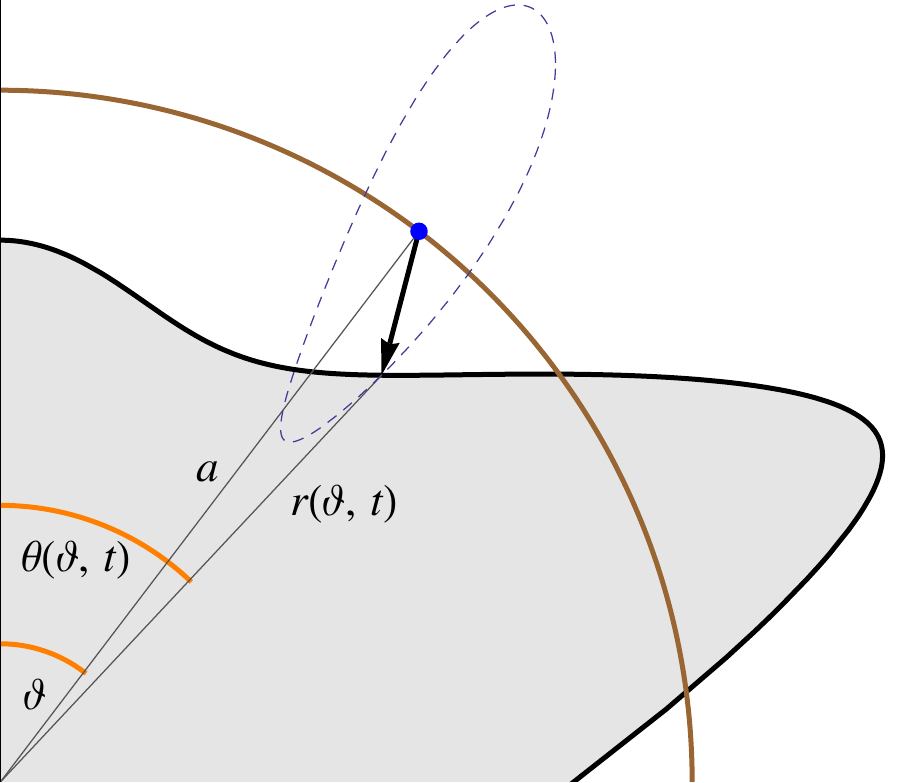}
\caption{Notation for axially symmetric distorted sphere, showing a cross section of a portion of the distorted sphere (gray) and undistorted surface (brown circle). At time $t$, the material point at radius $a$ and angle $\vartheta$ on the undistorted sphere is displaced as indicated by the arrow to location $r$, $\theta$. During an oscillation period, the material point moves around the dashed curve. 
The cases we consider have maximum distortion of a few percent of the radius $a$, so the distortion in this figure is exaggerated.}\figlabel{oscillating surface}
\end{figure}

For simplicity, we focus on small amplitude surface oscillations. Such oscillations are convenient for two reasons. First, piezoelectric materials can implement such oscillations. Second, small amplitude oscillations of simple geometries, such as a sphere, are analytically-tractable~\cite{happel83,blake71}. Neither of these reasons is necessary: metamorphic surfaces can give large shape changes~\cite{freitas99}, and numerical methods~\cite{michelin10} can evaluate locomotion for large amplitude deformations~\cite{shapere89}.

\subsection{Performance}

We focus on axially symmetric oscillations, which simplifies the performance analysis and gives the highest hydrodynamic efficiency~\cite{shapere89a}.
Due to viscous damping, fluid motion a distance $d$ from the surface is proportional to $e^{-d/\delta}$ where $\delta=\sqrt{2\viscosityKinematic/\omega}$ is the viscous damping length and $\viscosityKinematic=\viscosity/\density$ is the kinematic viscosity~\cite{fetter80}. 
When this damping length is large compared to the size of the robot, i.e.,  when $a/\delta \ll 1$, the flow is quasi-static: fluid response to each position of the oscillating surface is approximately the same as fluid would respond if the geometry were frozen in that configuration with boundary conditions determined by the shape and velocity of the surface at that time~\cite{kim05}.
Corrections to the quasi-static approximation are of order $a \sqrt{\omega/\viscosityKinematic}$, the Womersley number~\cite{kim05}. 
This approximation treats the fluid as incompressible, which requires that motions are slow compared to the speed of sound, i.e., $a \epsilon \omega \ll c$, which holds for the scenarios considered here.

For small distortions (i.e., $\epsilon \ll 1$), the fluid velocity and pressure around the sphere are proportional to $\epsilon \omega$, locomotion speed is proportional to $\epsilon^2 \omega$ and power is proportional to $\epsilon^2 \omega^2$~\cite{blake71,shapere89a}.
Comparing the scenarios of \tbl{scenarios} involves choices of oscillation amplitude and frequency, i.e., values for $\epsilon$ and $\omega$. The scenarios specify locomotion speed, which constrains the product $\epsilon \omega$ but not the individual values.
As a specific choice, we consider the same oscillation amplitude, i.e., the same value of $\epsilon$, in both scenarios and vary the oscillation frequency to give the specified  locomotion speeds. 
\tbl{oscillation parameters} gives these parameters for both scenarios, and shows the quasi-static approximation is reasonable.

\begin{table}
\centering
\begin{tabular}{llll}
scenario				&	&\scenario{low}		& \scenario{high} \\
oscillation frequency		& $f$				& $2\,\kHz$	& $0.02\,\kHz$\\
angular frequency		& $\omega=2\pi f$	& $1.3\times 10^4\,\radian/\second$	& $1.3\times 10^2\,\radian/\second$\\
viscous damping length	& $\delta=\sqrt{2\viscosityKinematic/\omega}$			& $13\,\micron$	& $13000\,\micron$\\
Womersley number		& $a \sqrt{\omega/\viscosityKinematic}$	& $0.1$	& $10^{-4}$\\
\end{tabular}
\caption{Oscillation frequency and criteria for quasi-static Stokes fluid flow using parameters of \tbl{scenarios}.}\tbllabel{oscillation parameters}
\end{table}

The remaining design question is the shape of the oscillations, i.e., choices for the magnitude and phase of the amplitude for each mode, given in \eq{amplitudes}.
Maximum hydrodynamic efficiency occurs in the limit of a large number of infinitesimally small wavelengths~\cite{shapere89a}.
However, there is only minor increase in efficiency when using wavelengths smaller than about $1/10$ the robot size, corresponding to modes with $n\approx 10$. 
Moreover, high modes have significantly different motions of nearby points on the surface, which require more rapid control and precise actuators. 
Thus practical propulsion with surface oscillations will not use the arbitrarily large modes that maximize hydrodynamic efficiency.

The optimal amplitudes have a simple form in the limit of large modes~\cite{shapere89a} and are close to optimum when applied to a finite set of modes $n=k,\ldots,k+p$. Specifically, these choices for the amplitudes in \eq{amplitudes} are
\begin{eqnarray}\eqlabel{optimal amplitudes}
A_n 			&=& (1+\sqrt{2}) \sin(j \psi) \\
B_n 			&=& \sin(j \psi) \\
\gamma_n	&=&	-\frac{\pi}{2}(j-1) \\
\eta_n		&=&	-\frac{\pi}{2}(j-1) \eqlabel{optimal amplitudes last}
\end{eqnarray}
with $j=n-k+1$, $\psi = \pi/(p+2)$, and otherwise $A_n=B_n=0$. The amplitudes multiply $\epsilon$ to produce the distortion of \eq{distorted sphere}, so have an arbitrary overall scale. We fix this scale by normalizing amplitudes so the largest displacement of any point on the sphere over the entire oscillation period is $a \epsilon$.

As an illustration, \fig{oscillating surface example} shows the surface distortions resulting from \eq{optimal amplitudes}--\eqbare{optimal amplitudes last} for $k=10$, $p=10$. The oscillations form waves traveling from the north to south pole, with the largest surface motions close to the equator. 
\tbl{oscillation performance} shows the performance of this locomotion. Even though hydrodynamic efficiency is low, the required power is well below that likely available to such robots.

\begin{figure}[th]
\centering \includegraphics[width=\figwidthS]{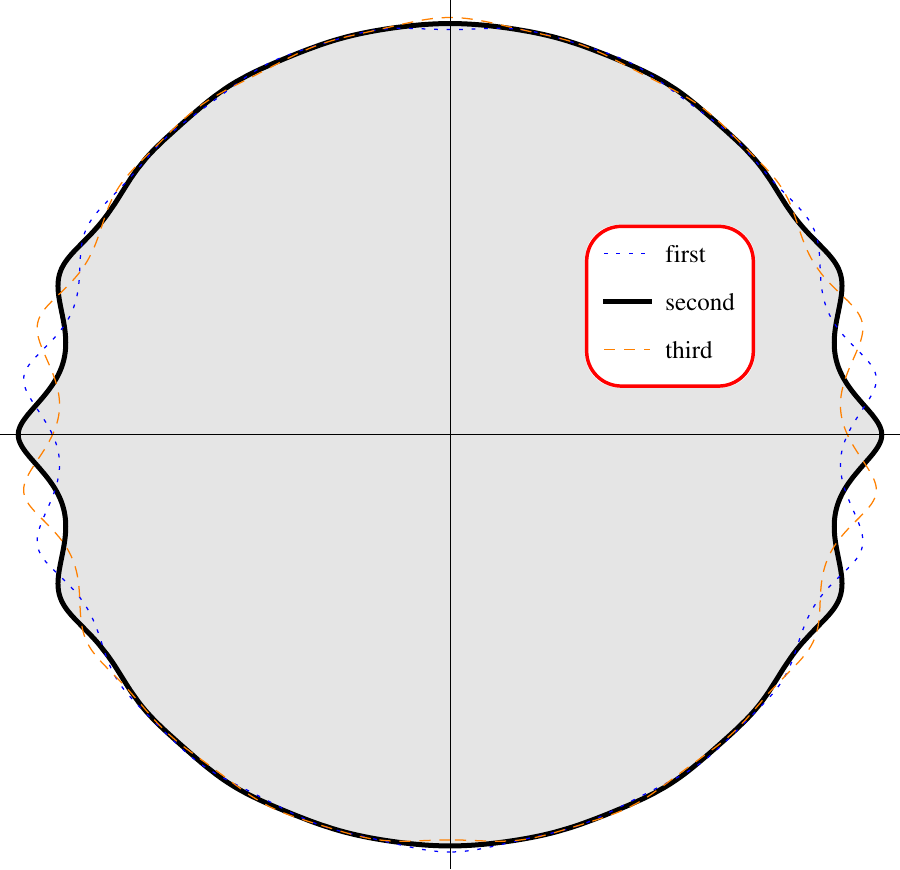}
\caption{Cross section of oscillating sphere at three equally spaced times during an oscillation period, in the order dotted (first), solid (second) and dashed (third). The gray region corresponds to the second position shown. Waves of surface distortion move from top to bottom, which is opposite the locomotion direction.\figlabel{oscillating surface example}}
\end{figure}

\begin{table}
\centering
\begin{tabular}{llll}
scenario				&	&\scenario{low}		& \scenario{high} \\
max. surface displacement	& $a \epsilon$					& $0.05\,\micron$	& $0.05\,\micron$\\
max. surface speed			& $a \epsilon \omega$			& $620\,\micron/\second$	& $6\,\micron/\second$\\
locomotion speed			& $3.29\, a \epsilon^2 \omega$		& $100\,\micron/\second$	& $1\,\micron/\second$\\
power				& $65.5\, a^3 \epsilon^2 \viscosity \omega^2$	& $0.025\,\picowatt$	& $0.025\,\picowatt$\\
hydrodynamic efficiency	& $3.12\,\epsilon^2$					& $0.008$		& $0.008$\\
max. thrust			& $62.1\,a^2 \epsilon^2 \viscosity \omega$	& $1.9\,\piconewton$	& $190\,\piconewton$\\
\end{tabular}
\caption{Performance of oscillating sphere using parameters of \tbls{scenarios} and \tblbare{oscillation parameters}, $\epsilon=0.05$ and amplitudes of \eq{optimal amplitudes} for $n=k,\ldots,k+p$, with $k=p=10$, normalized so maximum surface displacement is $a \epsilon$. The second column shows how the measures depend on geometry, oscillation amplitude and frequency, and fluid viscosity~\cite{blake71}.}\tbllabel{oscillation performance}
\end{table}

\begin{figure}[th]
\centering \includegraphics[width=\figwidth]{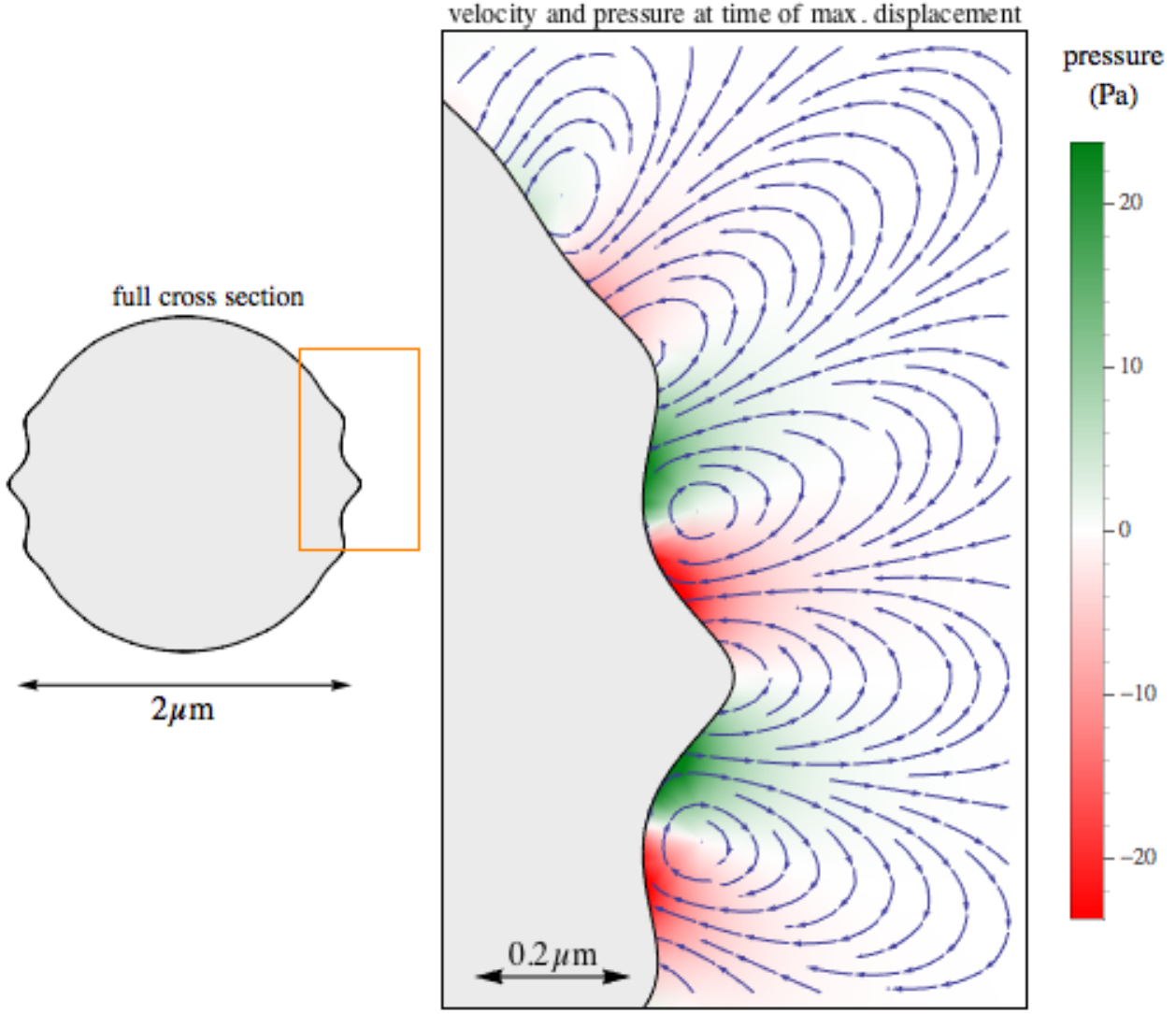}
\caption{Velocity streamlines and pressure in fluid around an oscillating sphere at the time during an oscillation period of maximum displacement for the \scenario{low} scenario. The parts of the surface surface with positive pressure, indicated in green, are moving outward, while parts with negative pressure are moving inward. The rectangular box on the full cross section at the left corresponds to the region shown in detail on the right.\figlabel{oscillating surface flow}}
\end{figure}

\fig{oscillating surface flow} shows the behavior of the fluid near the oscillating surface~\cite{blake71}. The magnitudes of the fluid velocity and pressure variation decrease rapidly away from the oscillating surface.

\subsection{Implementation}
\sectlabel{oscillation-implementation}

Microorganisms that move via coordinated waves of moving cilia produce fluid motion similar to small amplitude oscillations of the virtual surface formed by the outer envelope of those cilia~\cite{brennen77}.
Similarly, robots could use many closely-spaced short appendages to produce surface waves. However, using cilia reduces surface available for other devices, and cilia expose more surface area to the fluid which may be more likely to provoke immune reactions.
Instead, we consider two direct implementations of surface oscillations: electrically actuated piezoelectric materials and rods actuated by motors under a flexible surface. These are analogous to microorganisms moving via surface distortions produced by motion of internal structures~\cite{stone96b,ehlers11a,ehlers12}.

Alternatively, a mixed implementation could be useful with the piezoelectric material providing fine scale adjustments while portions of the surface requiring large oscillations (i.e., near the equator) use motors to produce most of the distortion.

\subsubsection{Piezoelectric actuators}

Piezoelectric materials change size in response to electric fields, with the change in length per volt of potential difference ranging up to $d=0.5\,\nanometer/\volt$ or so~\cite{uchino03}.
Piezoelectrics usually involve tiny motions, but some can produce strains of up to $5\%$~\cite{zhang11}, as used in the example of \fig{oscillating surface example}.
These materials are typically ceramics, although piezoelectric behavior also occurs in thin sheets such as graphene~\cite{ong12}.
The pressures on the robot from the fluid will have negligible effect on these materials. Specifically,
piezoelectric materials have Young's modulus of $E \sim 10^{10}\,\Pascal$ or more. Thus typical pressure changes in the fluid due to oscillations (e.g., tens of pascals, shown in \fig{oscillating surface flow}) give negligible changes in size.

The maximum displacement in the example of \fig{oscillating surface example} is $a \epsilon = 50\,\nanometer$, near the equator. Such displacements would require $a \epsilon/d \approx 100\,\volt$ across the material. Applied across an equatorial slab of material within a robot with $1\,\micron$ radius, this corresponds to an electric field of $10^8\,\volt/\meter$, at the upper end of fields encountered in microscopic biological contexts~\cite{freitas99}, though in this case the field is internal to the robot.
Such fields are considerably larger than usually applied to piezoelectrics, and thus may damage the material, e.g., by altering the polarization that gives rise to the piezoelectric effect.
Another difficulty in using piezoelectrics for the oscillation sizes considered here is their power dissipation, with dielectric losses typically 1\% of the energy involved in storing charge on the material~\cite{mezheritsky04}.

Thus, unless significantly more responsive (i.e., larger $d$ values), robust and low-dissipation piezoelectric materials become available, their use will be limited to smaller oscillations (i.e., smaller $\epsilon$) than the example in \fig{oscillating surface example}. This would give slower locomotion. Alternatively, using higher oscillation frequencies would achieve the same locomotion speed, but such combinations of smaller $\epsilon$ and larger $\omega$ would dissipate more power in the fluid, as shown in \tbl{oscillation performance}.

\subsubsection{Oscillating rods}

Mechanical forces on a deformable surface can produce surface oscillations.
For example, \fig{oscillation-internal} shows a rod displacing a portion of the surface at two different times, corresponding to maximum and minimum of the surface distance from the center of the robot. Motors~\cite{drexler92} could periodically move the rods to produce desired oscillation patterns~\cite{setter10}.

The rods indicated in the figure actuate the surface in one direction, i.e., radially, so any tangential motion would be a side effect of stretching the surface due to neighboring rods. A more complex rotary mechanism would be needed to produce a specific combination of radial and tangential motion illustrated in \fig{oscillating surface}. Alternatively, restricting oscillations to purely radial surface motion reduces locomotion speed compared to the optimal motions, which have both radial and tangential motions of \eq{optimal amplitudes}. For example, with no tangential motion (i.e., $B_n=0$), the locomotion speed is $60\%$ of the value in \tbl{oscillation performance}.

\begin{figure}
\centering
\includegraphics[width=\figwidth]{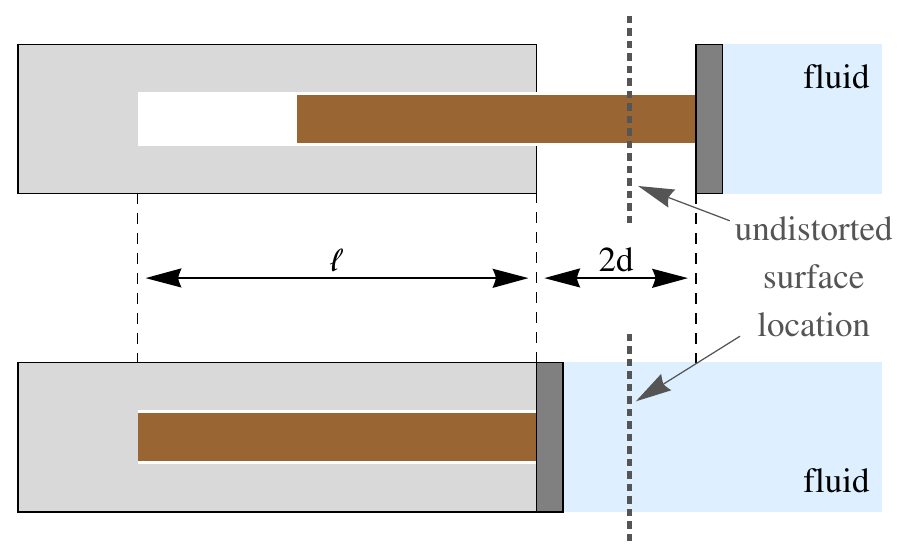}
\caption{Schematic of a rod moving a portion of the robot surface at the times of the oscillation period with maximum (top) and minimum (bottom) displacement. The motion range is twice the maximum displacement $d$ from the undistorted location (thick dotted line). The size of the gap between the rod and its housing is exaggerated. The segment of surface pushed by the rod is shown as a straight segment, but will actually bend to join with segments displaced differing distances by neighboring rods.}\figlabel{oscillation-internal}
\end{figure}

To produce the traveling waves on the surface, actuators must move regions of the surface corresponding to the highest modes with nonzero amplitude. The example of \tbl{oscillation performance} uses modes up to $k+p=20$. This requires independent motions at distances of $\delta=\pi a/20\approx 150\,\nanometer$, and hence about $4\pi a^2/\delta^2 \approx 500$ rods actuating the surface.

The main displacement is near the equator, leading to two implementation approaches. 
First, the robot could have rods uniformly spread across the surface. This would allow the robot to change direction by simply altering the oscillation pattern of the rods to correspond to the north-to-south axis oriented in the new direction.
Second, the robot could have actuators only in a band around its equator, thereby ignoring the small oscillations outside that band indicated by the amplitudes of \eq{optimal amplitudes}. Such an implementation would use fewer rods than the full implementation, but would need to rotate the robot to new directions, using non-axisymmetric oscillations analogous to those illustrated in \fig{rotation} for tangential motion.

As an example to estimate internal power use, consider 500 rods each with radius $r=50\,\nanometer$ and length $\ell=5d$, or $250\,\nanometer$, and the full circumferential area $S=2\pi r L$ sliding past the rod housing throughout the oscillation, i.e., we count the portion of the rod sticking out of the housing as also contributing to sliding friction.
This gives about $S=40\,\micron^2$ as the combined sliding surface area for all the rods.

The speed of the surface, and hence the rods, oscillates. Averaging \eq{friction} over an oscillation period $2\pi/\omega$ gives $\Pfriction = (1/2)k S v^2$ where $v$ is the maximum speed of the rod during the oscillation.
For an upper bound on the internal dissipation, suppose all the rods oscillate with this maximum displacement, so $v$ is the maximum surface speed $a \omega \epsilon$, given in \tbl{oscillation performance}.
These choices give $\Pfriction$ less than $8\times 10^{-3}\,\picowatt$ and $8 \times 10^{-7}\,\picowatt$ for the \scenario{low} and \scenario{high} scenarios, respectively. 
These values are smaller than the external power dissipation for both scenarios (\tbl{oscillation performance}), indicating internal dissipation is not significant for this implementation. Nevertheless, these values for $\Pfriction$ are larger than for tangential motion discussed in \sect{tangential-implementation}.

\section{Effects of Locomotion on Nearby Cells and Robots}
\sectlabel{safety}

\begin{figure}[th]
\centering \includegraphics[width=\figwidth]{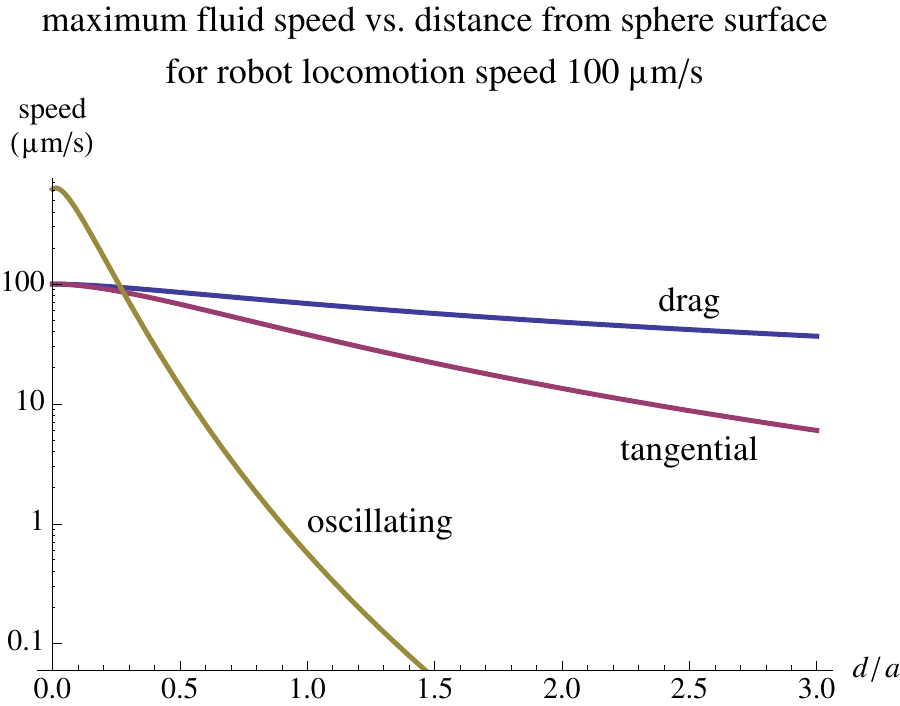} 
\caption{Maximum fluid speed due to a spherical robot, with $1\,\micron$ radius, vs.~distance $d$ from robot surface, relative to robot radius $a$, for different modes of locomotion in the \scenario{low} scenario. 
}\figlabel{fluid speed}
\end{figure}

Fluid motion can affect cells~\cite{davies95,discher05,papaioannou05,freitas03,vogel06}, including changing gene expression~\cite{chen01d}.
A minimal disturbance is the change in flow that activates mechanical sensors on the cell. For example, some cells can detect changes in velocity as little as $20\,\micron/\second$, shear (spatial gradients of velocity) in the range $1\mbox{--}10/\second$~\cite{guasto12}, and
fluid stress as low as $1\,\Pascal$~\cite{huang04}.

Such biological responses suggest conservative limits for microscopic robot locomotion are forces up to $10\,\piconewton$ and shear stresses up to $100\,\Pascal$~\cite{freitas03}. 
This range of stress is well below that required to significantly distort or rupture cell membranes~\cite{freitas99,nelson08}. 
The viscous stress from the fluid is $\viscosity \nabla v$. For the \scenario{low} scenario of \tbl{scenarios}, a shear of $1000/\second$ gives a stress of $1\,\Pascal$.

As an indication of the relative safety, \fig{fluid speed} illustrates maximum fluid speed produced by three ways of moving a robot: dragging by an external force, and the two locomotion methods discussed in this paper: steady tangential surface motion and small oscillations. 
Self-propelled robots disturb the fluid less than a robot dragged by an external force, which is also the case for microorganism movement~\cite{keller77}.
A change in velocity of at least $20\,\micron/\second$ occurs only within less than a micron from a sphere using tangential propulsion to move at $100\,\micron/\second$, but several microns for a sphere dragged by an external force. Using surface oscillations has larger effects close to the robot surface, but more rapid decrease with distance.

The gradient of the velocities discussed here indicate the shear and fluid stresses. For instance, in the \scenario{low} scenario, at a distance of $1\,\micron$ from the robot surface, the drag, tangential and oscillating modes of locomotion give shears of $30$, $40$ and $3/\second$, respectively. The corresponding fluid stresses are $0.03$, $0.04$ and $0.003\,\Pascal$, respectively.

For the \scenario{high} scenario, velocities and shears are 100 times smaller, while the viscosity is $10^4$ times larger, giving 100 times larger stresses at a given distance than in the \scenario{low} scenario.

In summary, the speeds and sizes considered here are not likely to immediately damage nearby cells. Less disruptive methods have higher safety margin. 
Low-disruption methods may be important for long-term use of the robots to reduce the possibility of undesired chronic response to shear forces~\cite{freitas03}.
For slow movement within cells, where the robot will pass internal structures at less than micron distances, the rapid decrease in disturbance with distance from the oscillating surface method could be particularly useful.
Conversely, robots could deliberately exert forces on cells by activating propulsion while blocked by the cell membranes. These forces could achieve pressures high enough to affect the cells, thereby imposing microscopically precise patterns of forces throughout tissues.

In addition to their effect on nearby cells, robot motion will affect other nearby robots~\cite{happel83,kim05}. Locomotion methods that reduce these interactions simplify navigation control for robots moving near each other.
One example is robots working together to build structures, especially with short time constraints and hence relatively rapid movements, such as forming clots in response to ruptured blood vessels~\cite{freitas00}. 
Another example is nearby communicating robots combining measurements to determine spatial gradients more accurately and more rapidly than individual robots~\cite{dusenbery98}.
For instance, one robot collecting chemical measurements from a particular location would benefit from minimal disturbance to the fluid from other passing robots.
For successive robots measuring spatial gradients of chemicals from nearby cells, reducing fluid disturbance each time a robot passes also reduces changes to the chemical gradient and hence reduces measurement noise. Thus less disruptive propulsion methods improve spatial gradient measurements repeated by multiple robots.

\section{Robot Shape}
\sectlabel{shape}

For simplicity, the above discussion focused on spherical robots. This section describes locomotion design constraints related to robot shape~\cite{dusenbery09,freitas99}.

\subsection{Geometric Constraints on Robot Shape}

Elongated shapes can have high hydrodynamic efficiency.
Nevertheless, design goals other than locomotion could significantly constrain the robot shape.
In particular, geometric constraints include:
\begin{itemize}
\item volume

The robot volume must encompass all required internal components, such as power generation, control and chemical storage, e.g., for drug delivery. 

\item surface area

The robot surface must be large enough for all components that interact directly with the robot's environment. These may include surface propulsion, sensors and pumps that collect chemicals from the surrounding fluid and components for forming mechanical connections with other robots.

\item minimum diameter

The minimum diameter determines the smallest gaps the robot could pass through, or the amount adjacent cells must be displaced when moving between them.

\item maximum diameter

The maximum diameter determines the minimum size vessels device can pass through if passing at random orientations or as a safety constraint if orientation control fails. 
The largest diameter also affects detection of chemical gradients using sensors located on opposite ends of the robot~\cite{dusenbery09}.

\item minimum radius of curvature

The minimum radius of curvature affects safety. A small radius means the robot has sharp ends that could puncture cells by applying the full force of locomotion over a tiny area.

The minimum radius of curvature also constrains the volume available for internal components next to the surface. 

Pointy ends give rise to large diffusive fluxes, which could overwhelm sensors or pumps even though those components readily handle the average flux over the entire surface. Conversely, pointy ends could be useful enhancements of diffusion of low-concentration chemicals to the robot as an alternative to using many sensors spread over the full surface of a less pointy shape.

\end{itemize}

These geometric constraints interact to restrict the allowable range of shapes, with some constraints dominant for near-spherical shapes while others, e.g., radius of curvature, dominating extremely elongated shapes.

\subsection{Example: Prolate Spheroids}

While most shapes are not amenable to analytic solutions, the results discussed above for spheres extend to spheroids~\cite{happel83,dusenbery09}. 
In particular, generalizations of  \eq{tangential motion} and \eqbare{tangential power} show that a prolate spheroid (\fig{prolate}) has larger hydrodynamic efficiency and less disturbance to the fluid than a sphere of the same volume~\cite{leshansky07}.
The small-amplitude analysis discussed above extends to spheroids and non-axisymmetric surface motions such as helical surface waves~\cite{ehlers11a}.

\begin{figure}[th]
\centering \includegraphics[width=\figwidthS]{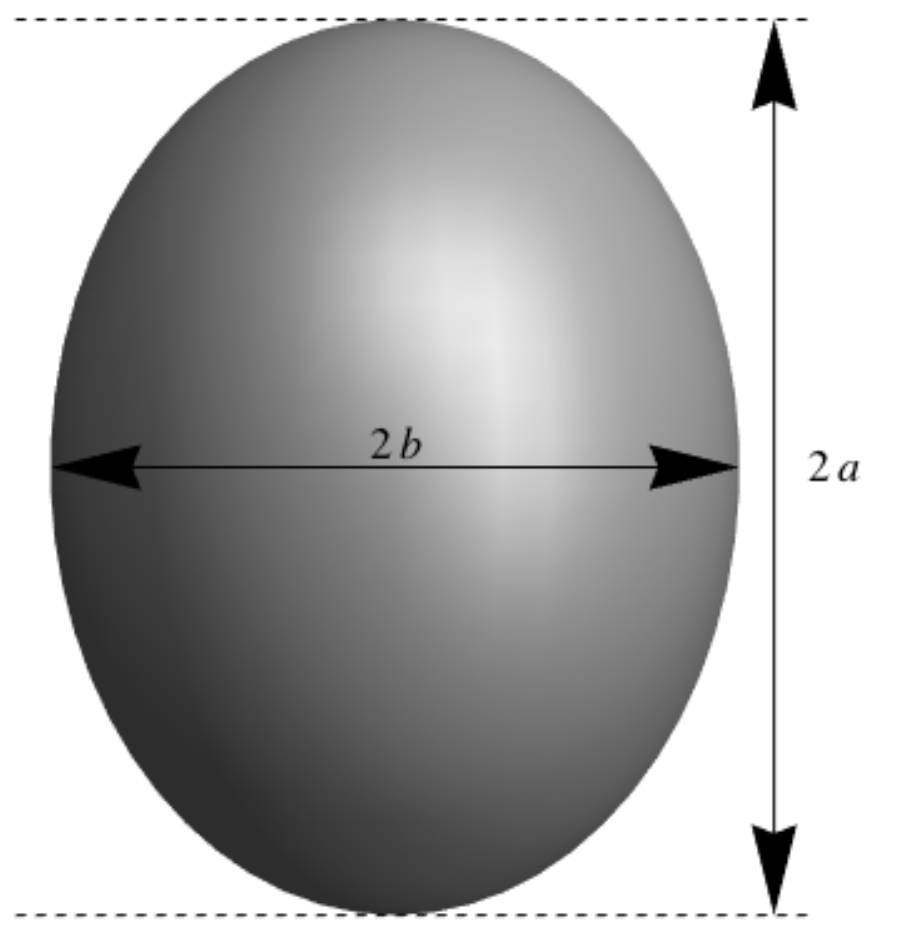} 
\caption{Prolate spheroid with semi-major and semi-minor axes of length $a$ and $b$, respectively, with $a> b$. When $a=b$, the shape is a sphere with radius $a$.}\figlabel{prolate}
\end{figure}

\begin{figure}[th]
\centering \includegraphics[width=\figwidth]{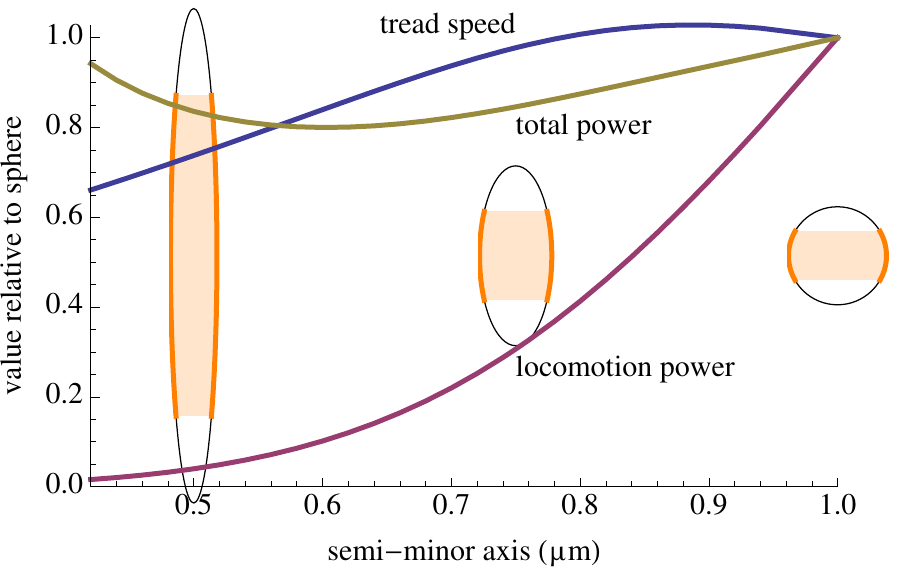}
\caption{Tread speed, locomotion power and total power, relative to values for a sphere, as a function of robot shape for the \scenario{low} scenario. Locomotion is from steady tangential motion on a band around the equator, as discussed in \sect{treadmill}. Robots move at the same speed ($U=100\,\micron/\second$), and have the same volume and surface area available for non-propulsion components. The ovals are cross sections of the robots corresponding to three values of the semi-minor axis indicated on the horizontal axis. The bands around the equatorial regions indicate the surface area devoted to propulsion.}\figlabel{shape}
\end{figure}

Comparing different shapes requires selecting appropriate geometric constraints. A common choice is to compare shapes with constant \emph{total} volume~\cite{dusenbery09}.
For robots, a more realistic choice accounts for the constraints required by the robot task, not just for locomotion itself. That is, locomotion gets the robot to suitable locations, but it needs other components to perform its task at the location. For example, the robot could need to carry a certain volume of drug to deliver to a cell identified by sensors on its surface. 
Thus as an example of design trade-offs related to shape, consider a task requiring the robot move at speed $U$ and have volume and surface area available for non-propulsion components of at least $\VnonpropulsionMin$ and $\SnonpropulsionMin$, respectively.
Propulsion components not only use some of the robot surface, but also occupy some volume of the robot near the surface. Thus both the total volume and surface area could vary with the shape.

As a specific case, we consider tangential motion on an equatorial band covering half the sphere's surface, as discussed with \tbl{tangential performance}, and implemented with treadmills with sizes given in \sect{tangential-implementation}. The tread housing extends a distance equal to about the bearing diameter $2r=100\,\nanometer$ below the robot surface. Thus propulsion covering surface area $\Spropulsion$ also uses volume $2r \Spropulsion$ within the robot.
As an example, suppose a sphere with radius $\aSphere=1\,\micron$ has half its surface area available for non-propulsion components, and this is the smallest radius satisfying both volume and surface area requirements. The geometry constraints are then
\begin{eqnarray}
\eqlabel{volume constraint}
\Vnonpropulsion &\geq& \VnonpropulsionMin = \frac{4\pi}{3}\aSphere^3 - 2 r (2\pi \aSphere^2) = 3.56 \micron^3\\
\eqlabel{surface constraint}
\Snonpropulsion &\geq& \SnonpropulsionMin = 2\pi \aSphere^2 = 6.28\micron^2 
\end{eqnarray}
An additional geometric constraint is that the robot is wide enough over the equatorial band to fit the tread housings. In the cases considered here, this constraint requires that the semi-minor axis $b$ of prolate spheroids satisfy $b\gtrsim 0.3\,\micron$.

Elongated shapes have more surface area for a given volume than a sphere. Thus a reasonable comparison is among shapes of the smallest possible volume and total power dissipation. 
Our estimate for total power is the sum of the locomotion power, $\Ppropel$, and the upper bound estimate for internal dissipation, $\Pinternal$, for the treadmill implementation described in \sect{tangential-implementation}.
This choice gives $\Vnonpropulsion=\VnonpropulsionMin$ and $\Snonpropulsion=\SnonpropulsionMin$, i.e., it is best to use all additional surface area from an elongated shape for propulsion. One reason for this choice is the increasing hydrodynamic efficiency with size of the band, analogous to the case for spheres seen in \tbl{tangential performance}.
Internal power dissipation is a more significant factor.
More propulsion components increase the sliding surface area but also allow the treads to operate more slowly for the required locomotion speed $U$.
In this case, the decrease in tread speed more than offsets the increase in sliding area, leading to minimum power use when the tread covers as large an area as allowed by \eq{surface constraint}.

\fig{shape} shows the performance of various shapes with these constraints.
The tread speed is relatively insensitive to shape, but hydrodynamic efficiency is significantly larger for elongated robots, resulting in lower locomotion power~\cite{leshansky07}. 
On the other hand, total power increases for sufficiently elongated shapes due to the increasing internal dissipation.

The increase in total power use for narrow robots illustrates the importance of including internal power dissipation in evaluating designs: the common focus solely on hydrodynamic efficiency would, in this case, misleadingly suggest using highly elongated ``needle'' shapes, at least as far as power use is concerned.

By contrast, in the \scenario{high} scenario, internal drag is negligible. The 100 times slower speed gives $10^4$ times less internal friction, while the locomotion power is the same due to the higher viscosity.

\section{Brownian Motion}
\sectlabel{brownian}

Microscopic robots not only face challenges of moving in viscous fluids, but also Brownian motion, which randomly changes both the robot's location and orientation.
An example scenario affected by Brownian motion is a group of robots released together by a larger device with specified initial orientations determined by this larger device with more sensor, computational and navigation capabilities than the released robots. For instance, the larger device could measure chemical gradients to identify the direction for the released robots. In this scenario, the robots are intended to proceed a short distance (e.g., a few cell diameters) in those initial directions to reach their operating locations in spaces too small for the larger device. These directions could be uniformly distributed so the robots examine all nearby cells for diagnosis.
Alternatively, the robots may be aimed in the same direction, e.g., to carry a large volume of drug to one specific cell.
This raises the question of the sophistication of the robot navigation systems required to reach these locations. 

The simplest navigation is dead reckoning: the robots continue in their initial directions until they reach their target locations.
Brownian motion limits the accuracy of this approach, thereby requiring more complex navigation methods, such as biased random walks if chemical gradients indicate the direction of the target location~\cite{berg93}.
Another navigation method is using reference signals from nearby implanted navigation nodes~\cite{freitas99}. Alternatively, stabilizing orientation with a gyroscope would require a substantial fraction of a micron-size robot's volume~\cite{freitas99}.
An alternative to more complex navigation for individual robots is to use a larger number of simpler robots so that enough robots reach the intended locations to complete the task.
Swarms are an extension of this approach in which each robot adjusts its direction based on the observed orientation of its neighbors~\cite{bonabeau99}.
Such adjustments reduce the effect of Brownian motion: the average direction of swarm will change more slowly than the direction of an individual robot.
While these are possible approaches to navigation, identifying cases where dead reckoning is sufficient will allow using simpler robots.

The diffusion constant of a sphere of radius $a$ in fluid with viscosity $\viscosity$ at temperature $T$ is
\begin{equation}
D = \frac{\BoltzmannConstant T}{6\pi a \viscosity}
\end{equation}
where $\BoltzmannConstant$ is the Boltzmann constant.
The change in position after time $t$ is of order $\sqrt{6 D t}$.
Brownian motion also affects robot orientation by changing orientation of a sphere's axis with time constant~\cite{dusenbery09} 
\begin{equation}\eqlabel{orientation time}
\tau= \frac{4 a^3 \pi \viscosity}{\BoltzmannConstant T}
\end{equation}
\tbl{Brownian motion} gives these values for the two scenarios, and the typical displacement during the time a robot moves $100\,\micron$. The orientation time is much longer than needed for the robot to actively change its orientation, as described with \tbl{tangential performance-rotation}. 
Thus these effects of  Brownian motion are fairly minor for motion over the times and distances of the scenarios of \tbl{scenarios}.

\begin{table}
\centering
\begin{tabular}{lcll}
scenario				&	&\scenario{low}		& \scenario{high} \\
translation diffusion coefficient 	& $D$		& $2\times 10^{-13}\,\ms$			& $2\times 10^{-17}\,\ms$\\
rms displacement after $d=100\,\micron$	&  $\sqrt{6 D d/U}$ & $1\,\micron$ 	& $0.1\,\micron$ \\
%
%
time constant for orientation loss	& $\tau$	& $3\,\second$	& $8\,\hour$ \\
travel distance during orientation time	& $U \tau$	& $300\,\micron$	& $30\,\millimeter$ \\
motile diffusion coefficient	& \Dmotile 	& $10^{-8}\,\ms$	& $10^{-8}\,\ms$ \\
\end{tabular}
\caption{Effects of Brownian motion on a sphere with radius of $1\,\micron$ for the scenarios of \tbl{scenarios}. For comparison, small molecules in water at body temperature have $D\approx 10^{-9}\,\ms$. 
}\tbllabel{Brownian motion}
\end{table}

For self-propelled objects, a more significant effect is change in robot orientation as the robot moves.
Over times large compared to $\tau$, random orientation changes lead to diffusive motion of self-propelled objects characterized by the motile diffusion coefficient $\Dmotile=\tau U^2/3$ where $U$ is the robot's speed~\cite{berg93,dusenbery09}. 
\tbl{Brownian motion} shows this diffusion is much more rapid than that due to translational diffusion.

On the other hand, for times comparable to or shorter than the orientation loss time $\tau$, locomotion remains predominantly in the original direction. This is the relevant case for motion over a few cell diameters, $100\,\micron$ or so, or in very viscous fluids. In this regime, the main effect of orientation loss is a limit on how far a robot can accurately navigate by dead reckoning.
The root-mean-square (rms) angle change is $ \alphaRMS=\sqrt{t/\tau}$ in time $t$, during which the robot moves distance $U t$.  
If the robot must reliably move distance $d$ with rms angle change at most $\alphaRMS$, then it must move at least as fast as $U=d/t=d/(\tau \alphaRMS^2)$. 
One consequence of this constraint is an increase in energy required for smaller $\tau$: power dissipation is proportional to $U^2$ while the time for the motion is $d/U$, giving energy use proportional to $U$. 
Thus expending more energy on locomotion extends the useful range of dead reckoning.

Robot geometry significantly affects Brownian motion. For instance, orientation time  (\eq{orientation time}) increases rapidly with size, so even somewhat larger robots can more effectively use dead reckoning.
Robot shape is also important.
\fig{speed Brownian motion} shows an example with the same geometry constraints as in \fig{shape} for an allowed change in orientation of $\alphaRMS=20^\circ$ rms over a distance $d=20\,\micron$. This gives a location error of order $d \sin \alphaRMS = 7\,\micron$, i.e., less than a typical cell diameter. 
Elongated robots maintain direction for a longer time, thereby allowing them to meet the navigation requirement with slower motion. 
Elongated robots also dissipate less energy, both internal and external, during the motion: the smaller dissipation of lower locomotion speeds more than compensates for the increased time to reach the location.
Thus choices of locomotion speed and robot shape can extend the applicability of dead reckoning navigation.

\begin{figure}[th]
\centering \includegraphics[width=\figwidth]{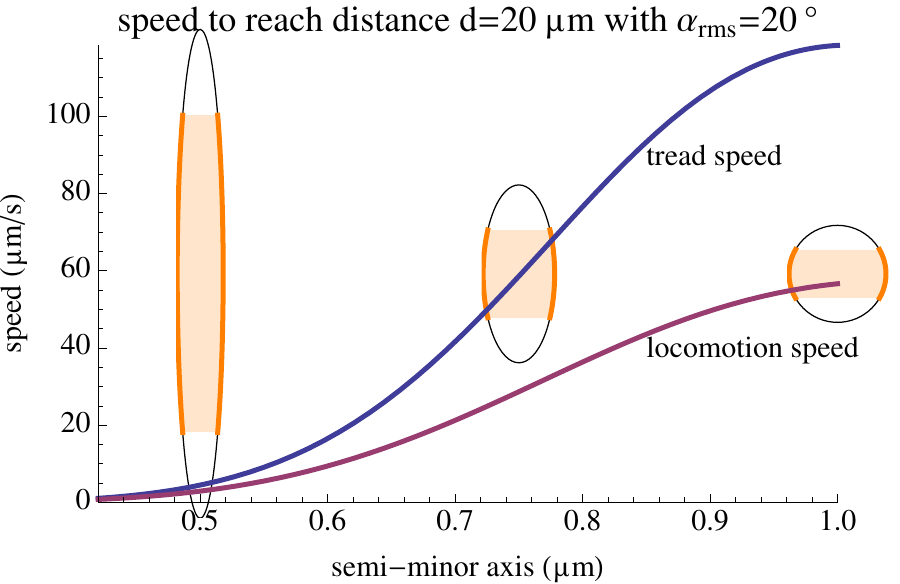} 
\caption{Locomotion speed required for dead reckoning navigation in fluid with the viscosity of the \scenario{low} scenario, over a distance of $20\,\micron$ for robots shaped as prolate spheroids with the same volume and surface area used for non-propulsion components. The ovals are cross sections of the robots corresponding to three values of the semi-major axis indicated on the horizontal axis. The bands around the equatorial regions indicate the surface area devoted to propulsion.}\figlabel{speed Brownian motion}
\end{figure}

\section{Choice of Propulsion Method}
\sectlabel{design choice}

This paper examines two propulsion mechanisms: steady tangential surface motion and periodic surface oscillations.
This section discusses their relative merits.

\subsection{Steady Tangential Motion}

Steady tangential surface motion can move the robot with large hydrodynamic efficiency and produce relatively high thrust and speeds compared to small amplitude oscillations. 

Tangential motion does not change the robot's shape. This could be useful when operating in regions with rigid walls where changing shape, especially change in volume, could result in large pressure changes in the surrounding fluid.

Treadmills could move the robot over solid surfaces~\cite{freitas99}  
in addition to movement in fluids, thereby providing operational flexibility.

In summary, propulsion by steady surface motion is especially useful for tasks that
1) require fast or efficient locomotion, especially in highly viscous fluids, 
2) operate in confined regions or alternate between moving in fluid and on solid surfaces, and
3) have significant available surface area to devote exclusively to propulsion.

\subsection{Surface Oscillations}

Forces applied within the robot can produce surface oscillations. Thus other components, such as sensors, can use the same surface area that provides motion (provided they function in spite of the oscillations). 
Since oscillation amplitude is largest near the equator, less flexible components could be placed near the poles.

At higher frequencies, oscillations can perform functions such as sensing, navigation and communication~\cite{freitas99,hogg12}. Thus an implementation that can operate at a wide range of frequencies can provide several functions.

With surface oscillations, a spherical robot can alter direction simply by changing the oscillation pattern. This contrasts with steady motion implemented by treadmills or wheels in fixed orientation on the surface, which requires a separate step to rotate the robot. 

Small-amplitude surface waves are likely easier to implement than methods requiring moving parts of the robot over long distances. In particular, oscillations avoid the need for watertight breaks in the surface.

Small-amplitude surface waves only slightly disturb surrounding fluid beyond a narrow boundary layer around the robot. 
Robots may need to enter and move through the interior of cells~\cite{freitas99} where propulsion methods with minimal disturbance may be particularly important to avoid damage. Such intracellular motion would be autonomous extensions of inserting probes into cells, which do not appear to damage the cells even after several days~\cite{gao12,xie13}.

In summary, propulsion by small surface oscillations is especially useful for tasks that
1) require most of the surface area for other uses and hence have little area available for separate propulsion components (without increasing the robot size), and
2) do not require fast or efficient locomotion.

\subsection{Example of Design Trade-offs}

Having presented two propulsion methods for microscopic robots, this section illustrates the choice among these methods in the context of conservative design constraints that provide large safety margins. Such choices are particularly appropriate for early use of robots with limited capabilities and which operate with significant uncertainties in the physical properties of their microenvironments and how cells respond to their motion.

We consider the following design constraints.
First, we allow up to $1\,\picowatt$ for power dissipation, leaving most of the robot's available power (i.e., tens of picowatts) for other tasks.
For shear stress, we use the conservative limit from \sect{safety} of $1\,\Pascal$ at a distance of $1\,\micron$ from the robot. 
We compute this shear force based on the gradient in fluid velocity near the robot surface shown in \fig{fluid speed}.
We suppose robots use dead reckoning for navigation, with the example discussed in \sect{brownian} as the limiting constraint.

\begin{figure}[th]
\centering \includegraphics[width=\figwidth]{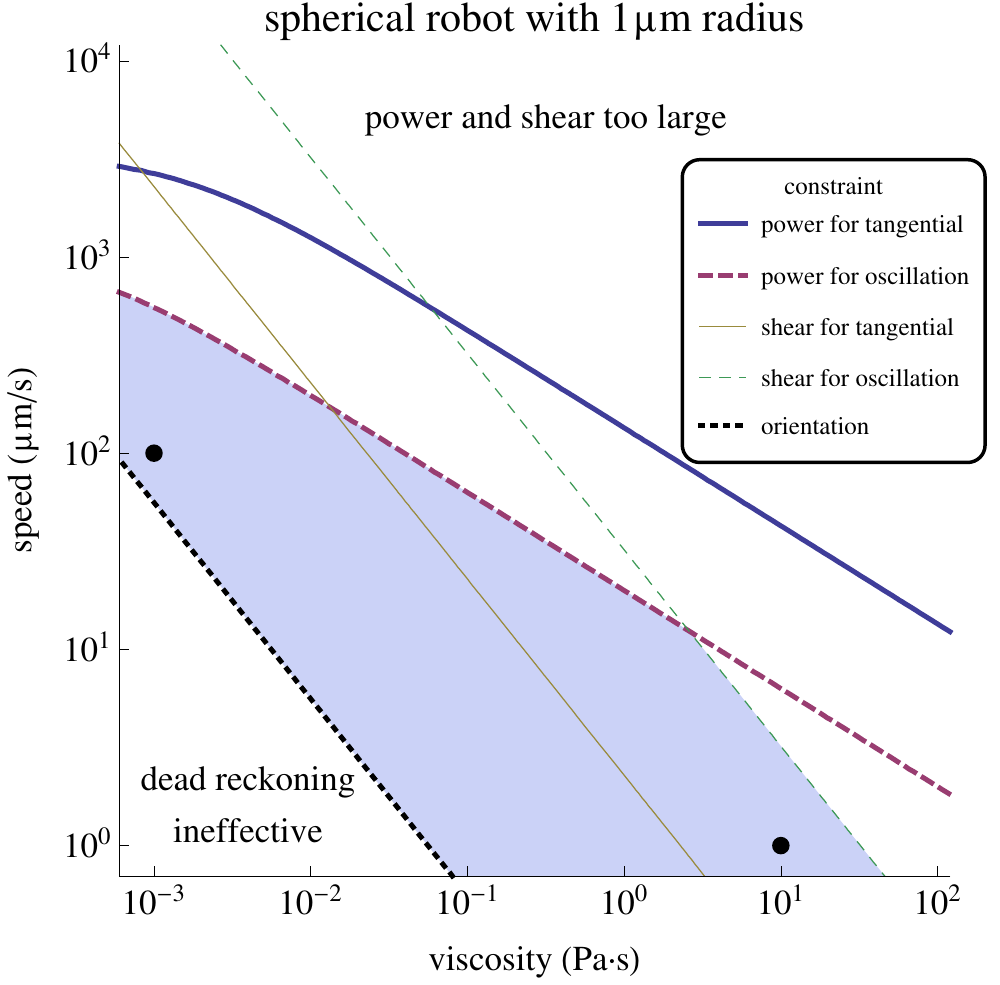} 
\caption{Performance constraints for a spherical robot using either tangential motion on an equatorial band covering half the surface (solid) or surface oscillations (dashed).
The axes use logarithmic scales.
The lines indicate constraints: power of $1\,\picowatt$, shear stress $1\,\micron$ from the robot surface of $1\,\Pascal$ and Brownian motion giving $\alphaRMS=20^\circ$ over a distance $d=20\,\micron$.
Surface oscillations satisfy all the constraints in the shaded area.
The black dots indicate the scenarios of \tbl{scenarios}.}\figlabel{tradeoffs}
\end{figure}

\fig{tradeoffs} illustrates design trade-offs from these constraints. The speed and viscosity values in the diagram cover much of the range relevant for \invivo\ operation.
The locomotion speeds range from $1\,\micron/\second$, allowing moving a few cell diameters in a minute, to $1\,\centimeter/\second$, a conservative upper limit to avoid damage to cells from collisions with the robots~\cite{freitas99}. 
The range of viscosities includes most biological fluids~\cite{freitas99},  
with water and blood plasma at the left, and cell cytoplasm on the right.

The diagram shows power and shear stress limit operation in high viscosity fluids or at high speeds\footnote{
At the upper left of the diagram, the Womersley number is close to 1 so the quasi-static approximation in \sect{oscillation} is less accurate~\cite{kim05}. 
Corrections are not significant for this example since this portion of the diagram is outside the feasible operation range.}. 
Tangential  motion is more efficient than surface oscillations, but also has larger shear stresses.
Brownian motion significantly affects navigation only for slower motion in low viscosity fluids.
Over most of the diagram, power limits arise mainly from external fluid drag. Only at the upper left, i.e., high speeds in less viscous fluids, do internal losses contribute noticeably.

Consequences of these limits depend on nanorobot task and operating environments. For instance, general purpose robots will operate at a wide range of speeds in fluids of various viscosities. Propulsion by surface oscillations provide a range of possibilities satisfying these constraints (shaded area in \fig{tradeoffs}).
On the other hand, specialized robots could be optimized for a limited range of speeds and fluid viscosity. One such example is robots that move rapidly in small blood vessels, especially to move upstream against the flow, requiring speeds around $10^3\,\micron/\second$ but only in low viscosity fluids. Another example is a robot needing to actively move only a few cell diameters away from capillaries over the course of several minutes, so speeds of $1\,\micron/\second$ or less are sufficient but the fluid may be highly viscous.

\fig{tradeoffs} also shows that tasks requiring higher performance (e.g., high speed in viscous fluids) will violate some of the constraints. For example, the robot could devote a more substantial fraction of its power to locomotion, thereby having less capability for other tasks, e.g., communication or on-board computation to evaluate sensor readings. 
Another example is moving slowly in low viscosity fluids which requires more complicated navigation than dead reckoning. 

Elongated robot shapes improve power efficiency and reduce the effect of Brownian motion, so are useful designs provided they satisfy geometric constraints, which are not included in \fig{tradeoffs}.
This discussion also does not include the maximum thrust force the robots could produce.

\section{Conclusion}
\sectlabel{conclusion}

This paper evaluated locomotion for microscopic robots in biological fluids.
Two appealing designs are steady tangential surface motion and periodic surface oscillations.
Both methods give speed and maneuverability sufficient for a variety of biomedical research, diagnostic and treatment applications. 
Moreover, these designs allow devoting a significant fraction of the robot surface to other, non-propulsion devices, such as sensors. These locomotion methods produce relatively little disturbance of the surrounding fluid and avoid potential damage or tangling from using extended structures such as flagella.
These features make locomotion compatible with other design goals, such as sensing, drug delivery and safety.
Furthermore, the machines will likely have more than enough power for these modes of locomotion.

Estimating internal power dissipation is challenging, both for microorganisms and robots, which leads most analyses of locomotion to focus on hydrodynamic efficiency. For microscopic robots made of precise, stiff materials, sliding friction dominates internal power dissipation, which allows rough estimates of internal dissipation. Including these estimates with the power evaluation emphasizes different designs than a focus on hydrodynamic efficiency alone would suggest. In particular, when considering robot shape, internal dissipation limits the benefit of increasingly elongated shapes.

A robot could have multiple locomotion modes for redundancy and operation in different environments. For instance, a robot could use treadmills for high-efficiency in fluids far from cells where damage to membranes is not an issue, and oscillating surfaces for locomotion near or within cells.

The implementations discussed in this paper focus on the surface components, which directly interact with the fluid, i.e., treadmills, wheels and oscillating surfaces. Robots need additional components to produce and control these surface motions. Thus a question for future study is designing and evaluating these components. One important property of these components is their energy dissipation. Another important property is their failure rates~\cite{drexler92}, which determine long-term reliability~\cite{tobias86} and hence the redundancy required to have high confidence that robots can complete their tasks.

Reliable locomotion of \invivo\ nanorobots require controls~\cite{freitas09} suited to the millisecond time and micron space scales of robot motion.
These controls include both individual robot behavior~\cite{alouges11,loheac13} 
and coordinated behaviors of multiple nearby robots (e.g., swarms) due to hydrodynamic interactions~\cite{kim05,riedel05,hernandez05,ishikawa08,kanevsky10,lauga09}. 
Sensors for the status of propulsion components can help ensure safe operation and aid locomotion control.
Such sensors include measuring the speed of propulsion components on the robot surface and forces on those components~\cite{freitas99}.
Unusual speeds or forces could indicate component failure or that the robot is stuck. The controller could respond to avoid damage to the robot or its surroundings by either fixing the problem (e.g., deactivating out-of-spec propulsion components) or, as a last resort, placing the robot in a passive ``safe mode''. 

A potential problem with rapidly moving components is whether they make the robot as a whole resonate, thereby building up large and potentially destructive oscillations.
As an estimate of relevant resonant frequencies,
consider a piston with stiffness $k_s \sim 25 \,\mbox{N/m}$ and mass $m  \sim 4 \times 10^{-15}\,\kg$ for a cubic micron nanorobot as typical of proposed materials for microscopic robots. The lowest resonant frequency is $f_{\mbox{\scriptsize res}} \sim 10 \,\MHz$~\cite{freitas99}.  
This is much larger than the propulsion frequencies we consider, so resonances from locomotion in micron-scale robots are unlikely.

Fabrication of complex structures with stiff materials may be able to create other propulsion methods, including jets with the possibility of arbitrarily small disturbance to the surrounding fluid~\cite{spagnolie10}, propellers~\cite{wang07a},
acoustic streaming~\cite{lighthill78,ehlers11}, and moving extended structures such as flagella and cilia. For example, flagella consisting of stiff, telescoping rods could reduce drag,  compared to similar microorganism propulsion, by retracting the rod during the recovery stroke, and simultaneously reduce the opportunities for tangling or damaging nearby cells.

The analyses in this paper focus on steady locomotion in simple, unbounded fluids with fixed properties on the scale of the robots.  Only variation due to large differences in fluid viscosity are considered in this paper.
Important extensions beyond this setting include motion in biological tissues with significant viscoelasticity~\cite{lauga09,freitas99}, 
and motion through small spaces, e.g., through narrow channels within bone, where boundary effects are significant~\cite{happel83,kim05,hernandez05,riedel05}. 
Moreover, the viscosity of some biological fluids can vary significantly with the size and surface coatings of objects in those fluids~\cite{lai09}.
Robots may be able to exploit these effects to improve locomotion by, for example, harnessing interactions with nearby boundaries~\cite{trouilloud08} or tuning size and surface properties to minimize viscosity as proposed for engineering nanoparticle-based drug delivery~\cite{lai09}.
Another extension is moving in bursts, which could use more power than available for steady motion, hence achieving higher speeds or forces. This could be especially useful for operating in very viscous fluids.

Biological tissues vary at both large and small space and time scales in response to environmental signals. Feedback control can allow robots to compensate for such changes. Moreover, robots could initiate such changes through active signaling to alter the robot's environment as it moves. For example, the propulsion mechanism can apply mechanical forces on individual nearby cells. Alternatively, the robot could release chemical signals from onboard storage tanks.
A potential application for such signals is moving between cells forming the boundary of tissues, such as blood vessel walls, without damaging cells, by signaling them to change adhesive forces, similar to signals used by white blood cells to exit vessels~\cite{ager03,ley07}.

Robots could act together to produce larger mechanical forces or chemical concentrations than a single robot. Such coordinated activity includes creating patterns of forces on cells over extended periods of time, which can change the cell arrangement and function~\cite{davies95}.

Future evaluation of these questions will clarify locomotion design trade-offs for microscopic robots. 
Fabrication of such robots involve significant technological challenges. Prior to the feasibility of such fabrication, theoretical studies, such as presented in this paper, quantify likely nanorobot capabilities and their suitability for various applications.
In particular, the evaluation of locomotion options provides guidelines for designs and indicates alternative implementations providing the same capabilities. This variety of options suggests microscopic robots will have several feasible locomotion methods.


\end{document}